\documentclass{article}

\usepackage[preprint]{neurips_2026}

\usepackage[utf8]{inputenc} 
\usepackage[T1]{fontenc}    
\usepackage[hidelinks]{hyperref}       
\usepackage{url}            
\usepackage{booktabs}       
\usepackage{amsfonts}       
\usepackage{nicefrac}       
\usepackage{microtype}      
\usepackage{xcolor}         
\usepackage{makecell}
\usepackage{graphicx}
\usepackage{subcaption}
\usepackage[accsupp]{axessibility}
\usepackage[most]{tcolorbox}
\usepackage{wrapfig}
\usepackage{longtable}
\usepackage{array}
\usepackage{wrapfig}
\usepackage{amsmath}
\usepackage{amssymb}
\usepackage{mathtools}
\usepackage{amsthm}
\usepackage{alltt}
\newcommand{\EMA}[1]{{\color{red} EMA: #1}}

\usepackage[utf8]{inputenc}
\usepackage{newunicodechar}
\newunicodechar{ϵ}{\ensuremath{\epsilon}}

\newcommand{\BENCH}{\textit{OrigamiBench}}
\title{Can AI Understand  the Language of Origami?}

\author{Naaisha Agarwal\thanks{Shared first author}\\
Independent Researcher\\
{\tt\small naaishaagarwal@gmail.com}
\and
\textbf{Yihan Wu}\footnotemark[1]\quad \textbf{Xin Guan}\footnotemark[1] \quad \textbf{Yao-Xiang Ding}\\
State Key Lab of CAD\&CG, Zhejiang University\\
{\tt\small \{wu\_yihan, 3230105741,yxding\}@zju.edu.cn}\\
\and
\textbf{Ayaan Garg} \\
Independent Researcher\\
{\tt\small gargayaan10@gmail.com}
\and
\textbf{Vincenzo Collura} \\
SnT, University of Luxembourg\\
{\tt\small vincenzo.collura@uni.lu}\\
\and
\textbf{Yikuan Hu}\quad \textbf{Mohan Li}\quad \textbf{Wang-Zhou Dai} \\
National Key Laboratory for Novel Software Technology,
Nanjing University\\
{\tt\small \{huyk,limh,daiwz\}@lamda.nju.edu.cn}\\
\and
\textbf{Emanuele Sansone}\thanks{Last author} \\
CSAIL, MIT / ESAT, KU Leuven \\
{\tt\small esansone@mit.edu}
}

\begin{document}

\maketitle

\begin{abstract}
Building AI systems that can plan, act, and create in the physical world requires more than pattern recognition. Such systems must reason about the generative mechanisms and constraints governing physical processes, using structured representations that connect observations, actions, and their effects. Yet, many existing benchmarks study these capabilities separately, focusing either on visual recognition or on abstract symbolic or programmatic reasoning. Origami provides a natural testbed that integrates these abilities: constructing shapes through folds requires visual perception, reasoning about geometric and physical constraints, and sequential planning, while remaining sufficiently structured for systematic evaluation. We introduce \BENCH{}, a benchmark for evaluating programmatic understanding of the mechanisms underlying origami synthesis through a high-level language of physically grounded fold actions. Experiments with modern vision-language models reveal that scaling model size alone does not reliably improve reasoning about physical transformations. Moreover, models struggle to ground programmatic information in visual observations, suggesting that visual and language representations remain weakly integrated.

\end{abstract}


{\section{Introduction}\label{sec:intro}}
Building AI systems that can plan, act, and create in the physical world requires more than pattern recognition. Such systems must reason about the generative mechanisms and constraints governing physical processes in order to make sequential decisions. This requires structured representations that connect observations, actions, and their effects on the environment, while capturing how complex outcomes arise from sequences of simpler transformations. Yet, many existing benchmarks study these capabilities separately, focusing either on recognizing visual states or on reasoning over abstract symbolic or programmatic representations. This raises a fundamental question: \emph{can models learn to reason about how physical outcomes are generated through structured representations that connect observations, actions, and their effects?}

Origami provides a natural setting in which perception, physical validity, and symbolic composition can be studied jointly. A sheet of paper is transformed through a sequence of folds constrained by geometry and foldability~\citep{akitaya2013generating}. Despite relying on a single primitive action, repeated folding generates a rich space of designs through an inherently compositional and recursive synthesis process~\citep{gibbons2003origami}. Existing computational approaches, however, predominantly represent origami through crease patterns~\citep{lang1996computational,tachi2010freeform,tachi2009origamizing,akitaya2013generating,zahavy2026corigami,huang2026learn2fold}, which describe the geometry of a design but do not explicitly encode the ordered composition of folds that produces it. They therefore make the \emph{outcome} of synthesis explicit while leaving the underlying \emph{sequential generative mechanism} implicit.

In this work, we make the synthesis process itself the object of representation and reasoning. To this end, we introduce \BENCH{}, a benchmark for quantifying how well modern vision-language foundation models (VLMs) understand the mechanisms underlying origami synthesis. At its core is a high-level language in which origami designs are expressed as compositions of physically grounded fold actions, bridging visual states with the sequential transformations that generate them. The language abstracts away low-level geometric implementation of crease patterns while preserving the structure required to reason about ordering, composition, and physical validity. This enables us to precisely annotate folding trajectories, generate physically valid synthesis problems with controllable compositional complexity, and construct evaluation tasks that probe whether models can ground symbolic transformations in visual observations. Experiments on \BENCH{} reveal three key findings: scaling model size alone does not reliably improve reasoning about physical transformations; models struggle to ground structured programmatic information in visual observations; and this difficulty is particularly pronounced for fine-grained program elements such as fold points, fold types, and paper layers. Overall, these results suggest that visual and language representations remain weakly integrated.

{\section{OrigamiBench}\label{sec:bench}}
\begin{figure*}[t]
\vspace{0.5cm}
\begin{center}

\includegraphics[width=\textwidth, trim=0.cm 1.cm 0.cm 6.cm]{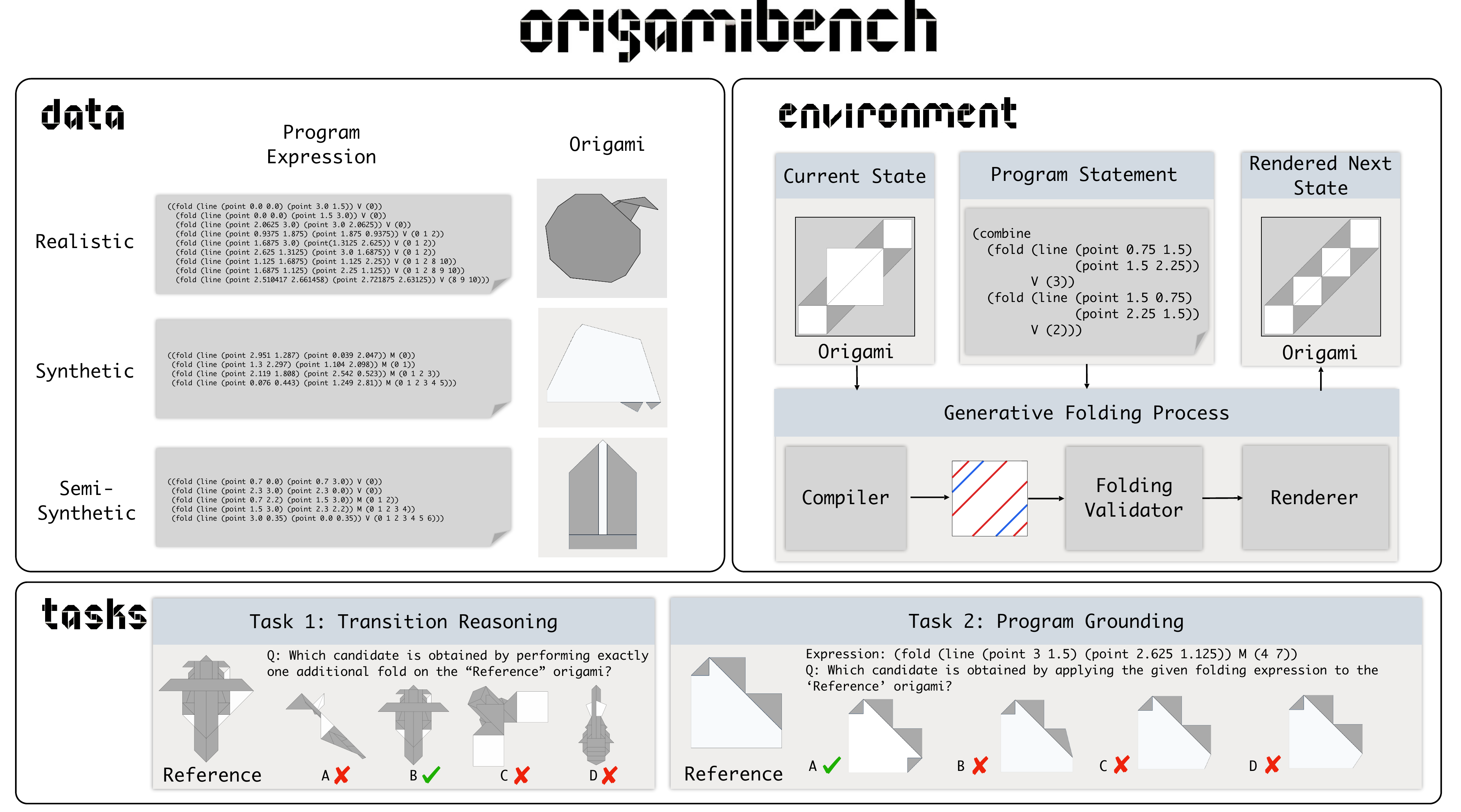}
\end{center}
\caption{Visual summary of \BENCH{}. 
On the left (\textbf{Data}), representative realistic, synthetic, and semi-synthetic examples are shown as program expressions paired with their rendered origami states. On the top right (\textbf{Environment}), a program statement is executed from the current state through the generative folding process: the compiler converts the statement into a crease-pattern representation, the folding validator checks the resulting state, and the renderer produces the rendered next state. At the bottom (\textbf{Tasks}), the two multiple-choice evaluations are illustrated from the model's perspective. Task 1 (\textbf{Transition Reasoning}) provides the current state and candidate successors, whereas Task 2 (\textbf{Program Grounding}) additionally provides the generating program statement. Green and red marks indicate the correct and incorrect states respectively.
}
\label{fig:origamibench}
\end{figure*}

\subsection{Origami Action Language}
\label{origami_grammar}
We represent an origami design as a program consisting of a sequence of folding actions applied to a paper state. Unlike geometric representations such as FOLD~\citep{demaine2016new}, which encode crease patterns and folded states, our language specifies the transformations that produce them. Programs consist of atomic folds, compound folds, paper turns, and reusable folding primitives, and can be compiled into FOLD graphs for foldability checking and rendering. The syntax is defined in Figure~\ref{fig:grammar}.

\begin{figure}[htbp]
\centering
\begin{verbatim}
<program>    ::= (<statement>*)
<statement>  ::= <fold> | <combine> | <library> | (turn h) | (turn v)
<fold>       ::= (fold <line> <fold_type> <levels>)
<line>       ::= (line <point> <point>)
<point>      ::= (point <coordinate> <coordinate>)
<coordinate> ::= [-]<number>
<number>     ::= <integer> | <digit>+.<digit>+
<integer>    ::= <digit>+
<digit>      ::= 0 | 1 | 2 | 3 | 4 | 5 | 6 | 7 | 8 | 9
<fold_type>  ::= M | V
<levels>     ::= (<integer>+)
<combine>    ::= (combine <fold> <fold>+)
<library>    ::= (<signature> <args>)
<signature>  ::= pleat | crimp | inrev | outrev | rabear
<args>       ::= (<point>+ <levels>)
\end{verbatim}
\caption{Grammar of the origami action language. Programs are Lisp-like expressions composed of atomic folds, compound folds, library operations, and paper turns.}
\label{fig:grammar}
\end{figure}
A \texttt{program} is an ordered sequence of \texttt{statement}s that transform the current paper state. Statements are evaluated sequentially and may be atomic \texttt{fold}s, compound \texttt{combine} operations, calls to reusable \texttt{library} operations, or turns of the paper about its horizontal \texttt{(turn h)} or vertical \texttt{(turn v)} axes.

\begin{figure*}[t!]
\begin{center}
\includegraphics[width=\textwidth, trim=0.cm 2.cm 0.cm 2.cm]{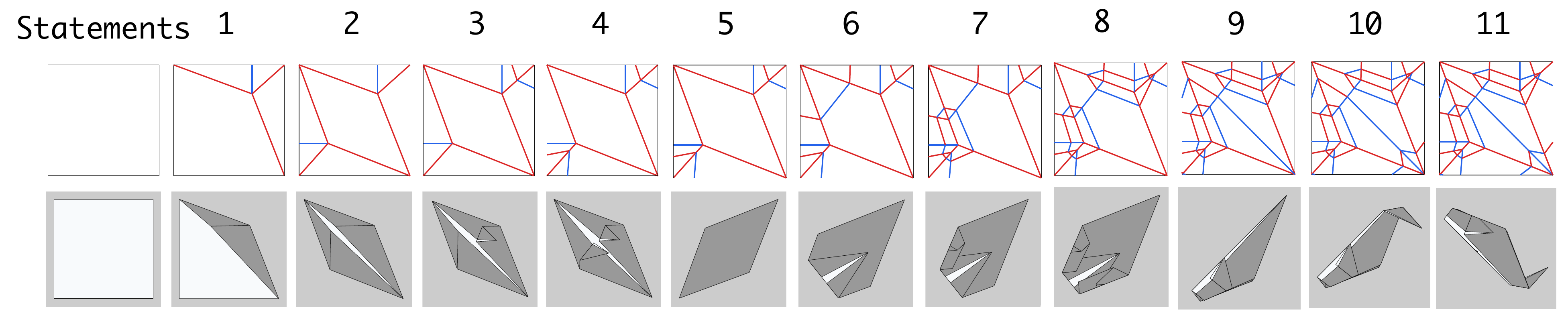}
\end{center}
\caption{Execution of an origami action program that constructs a whale. Each column corresponds to one program statement. The statement is compiled into a crease graph (middle row), checked for foldability, and executed to obtain the updated rendered paper state (bottom row). Thus, statements in the language correspond directly to folding actions rather than individual crease edges. In particular, a single atomic action may induce several crease edges: statement 6, for example, produces two valley and one mountain crease.}
\label{fig:example}
\end{figure*}

An atomic \texttt{fold} is parameterized by a line, a fold direction, and the paper layers affected by the operation. The line is specified by two 2D points, and the fold direction is either mountain (\texttt{M}) or valley (\texttt{V}). The \texttt{levels} argument identifies the indexed paper layers participating in the fold. Explicit layer selection is necessary to create detailed, complex origamis, where folds are not applied arbitrarily to the entire origami but only to specific layers to create specific shapes. Layer indices are updated after each folding operation to reflect the current layers in the origami. Importantly, an atomic action does not necessarily correspond to a single crease edge: depending on the current paper state, one fold action may induce several mountain and valley crease edges (Figure~\ref{fig:example}).

A \texttt{combine} expression groups folds that are evaluated against the same paper state. It may differ semantically from executing the folds sequentially, where each fold acts on the state produced by the previous one. This distinction is necessary for folding actions whose constituent creases must be formed together and cannot be represented as a sequence of independent folds (as the case od some libraries).

A \texttt{library} expression represents a reusable folding primitive, such as a pleat, crimp, or reverse fold. Each primitive expands deterministically into base fold and \texttt{combine} actions according to Table~\ref{tab:libraries}. Library operations thus provide a compact hierarchical representation without extending the semantics of the base language. The number of point arguments is determined by the primitive signature in Table~\ref{tab:libraries}.

The \texttt{turn} statements reorient the current paper state about its horizontal (\texttt{turn h}) or vertical (\texttt{turn v}) axis before subsequent actions are applied.

\begin{table}[b]
\centering
\caption{Library primitives and their expansion into base folding actions.}
\label{tab:libraries}
\begin{tabular}{rlll}
\hline
\textbf{Fold} & \textbf{Name} & \textbf{Invocation} & \textbf{Expansion} \\
\hline
\raisebox{-0.3\height}{\includegraphics[width=0.6cm]{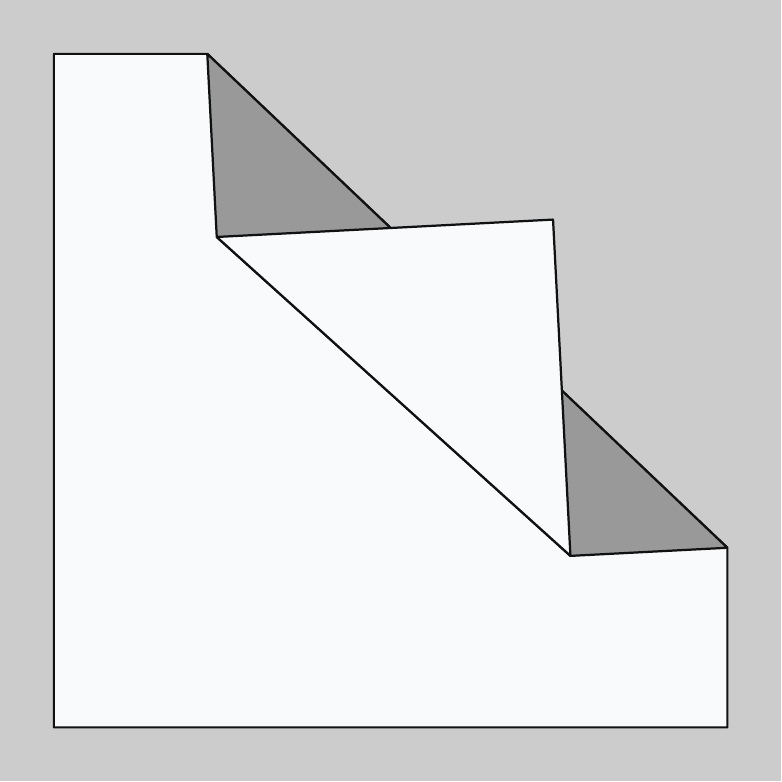}} & Pleat & \texttt{(pleat (X Y U W L))} & \makecell[l]{\texttt{(combine (fold (line X Y) V (L))} \\ \texttt{\phantom{ combine }(fold (line U W) M (L)))}} \\
\hline
\raisebox{-0.3\height}{\includegraphics[width=0.6cm]{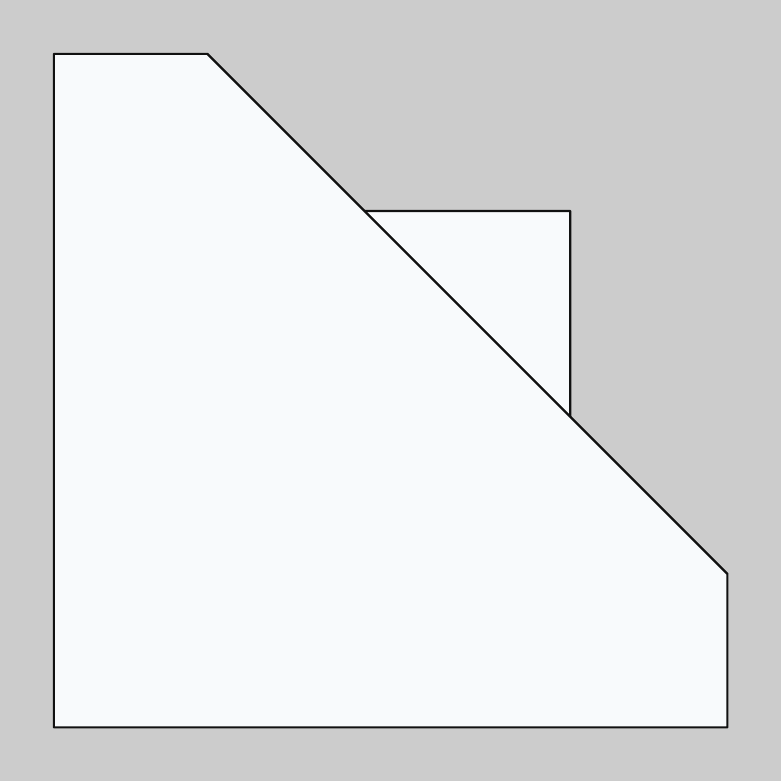}} & Crimp & \texttt{(crimp (X Y U W L))} & \makecell[l]{\texttt{(combine (fold (line X Y) M (L))} \\ \texttt{\phantom{ combine }(fold (line U W) V (L)))}} \\
\hline
\raisebox{-0.3\height}{\includegraphics[width=0.6cm]{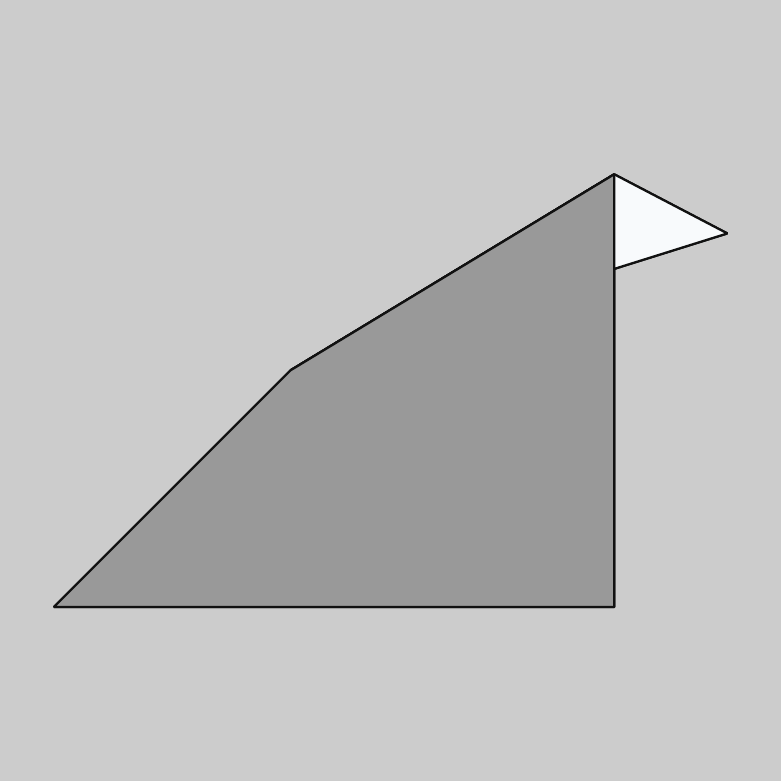}} & \makecell[l]{Inside \\ Reverse} & \texttt{(inrev (X Y U L N))} & \makecell[l]{\texttt{(combine (fold (line X Y) V (L))} \\ \texttt{\phantom{ combine }(fold (line X Y) V (N))} \\
\texttt{\phantom{ combine }(fold (line X U) M (L)))}} \\
\hline
\raisebox{-0.3\height}{\includegraphics[width=0.6cm]{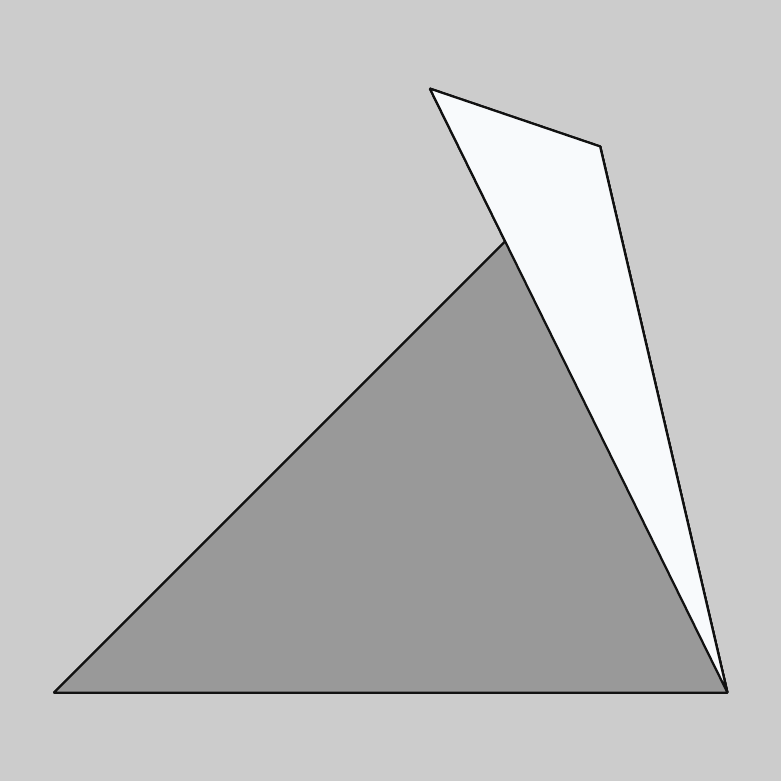}} & \makecell[l]{Outside \\ Reverse} & \texttt{(outrev (X Y U L N))} & \makecell[l]{\texttt{(combine (fold (line X Y) M (L))} \\ \texttt{\phantom{ combine }(fold (line X Y) M (N))} \\
\texttt{\phantom{ combine }(fold (line X U) V (L)))}} \\

\hline
\raisebox{-0.3\height}{\includegraphics[width=0.6cm]{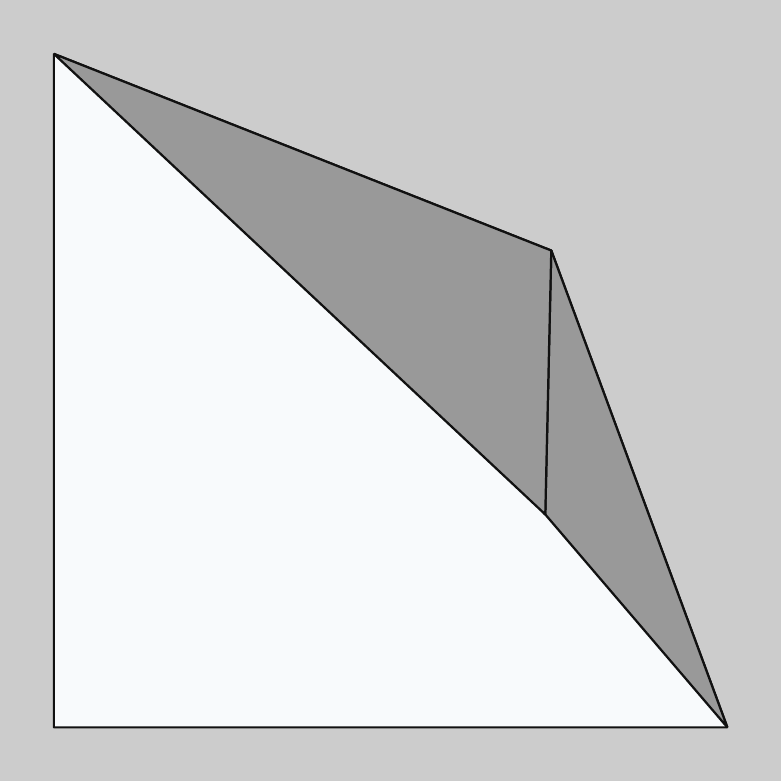}} & \makecell[l]{Rabbit \\ Ear} & \texttt{(rabear (X Y U W Z L))} & \makecell[l]{
\texttt{(combine (fold (line X Y) V (L))} \\
\texttt{\phantom{ combine }(fold (line X U) V (L))} \\
\texttt{\phantom{ combine }(fold (line X W) V (L))} \\
\texttt{\phantom{ combine }(fold (line X Z) M (L)))}
} \\
\hline
\end{tabular}
\end{table}

\textbf{Example.} Figure~\ref{fig:example} shows an eleven-statement program for constructing a whale. The first two statements are libraries, followed by a sequence of valley folds and turn operations. The complete program is shown below:\footnote{where coordinates are rounded to the first decimal for simplicity of visualization.}

\begin{verbatim}
(
    (rabear ((point 2.0 2.1) (point 0.2 2.8) (point 2.8 2.8) 
             (point 2.8 0.2) (point 2.0 2.8) (0)))
    (rabear ((point 0.9 1.0) (point 0.2 2.8) (point 0.2 0.2)
             (point 2.8 0.2) (point 0.2 1.0) (0)))
    (fold (line (point 1.3 1.9) (point 1.4 2.1)) V (2 3))
    (fold (line (point 1.2 1.7) (point 0.9 1.4)) V (5 6))
    (turn v)
    (fold (line (point 1.2 0.6) (point 0.6 1.2)) V (0 3 8))
    (fold (line (point 0.3 0.8) (point 0.8 2.1)) V (7 8 9 10 11 13))
    (fold (line (point 2.0 0.8) (point 0.8 0.3)) V (2 3 4 10 11 12 13))
    (fold (line (point 0.5 0.5) (point 2.8 2.8)) V (9 10 11 13))
    (fold (line (point 2.1 2.3) (point 2.3 2.5)) V (4 7 25 26))
    (turn h)
)
\end{verbatim}

\subsection{Environment Workflow}
We implement an environment that executes programs written in the origami action language introduced in Section~\ref{origami_grammar}. Starting from an unfolded sheet, the environment processes each program statement sequentially, updating the paper state accordingly. Each statement undergoes three stages: compilation, foldability checking, and rendering, as shown in Figure~\ref{fig:origamibench}.

\noindent\textbf{Compilation.}
The paper state is represented internally as a FOLD graph~\citep{demaine2016new}. The compiler, implemented using Rabbit Ear,\footnote{\url{https://github.com/rabbit-ear/rabbit-ear}} maps each statement and its corresponding paper state to a candidate successor graph. Atomic folds are resolved with respect to the selected layers and may introduce multiple crease edges when the fold line intersects several faces or overlapping layers (Figure~\ref{fig:example}). For \texttt{combine} operations, all constituent folds are compiled against the same pre-action state. Library primitives are first expanded according to Table~\ref{tab:libraries}. \texttt{turn} operations update the paper orientation and layer-index mapping.

\noindent\textbf{Foldability checking.}
The candidate FOLD graph is passed to Flat-Folder,\footnote{\url{https://github.com/origamimagiro/flat-folder}} which checks local flat-foldability conditions, including Maekawa's and Kawasaki's theorems, and then searches for a globally consistent folded configuration. If no feasible configuration is found, the action is marked invalid and the state remains unchanged; otherwise, the candidate graph becomes the new state.

\noindent\textbf{Rendering.} For a feasible statement, the validated folded configuration is rendered to produce the updated folded-paper view and the corresponding 2D crease pattern.

\subsection{Data Generation and Organization}

\begin{wrapfigure}{r}{0.4\textwidth}
    \centering
    
    \includegraphics[
        width=0.8\linewidth,
        trim=0cm 0.5cm 0cm 0.5cm,
        clip
    ]{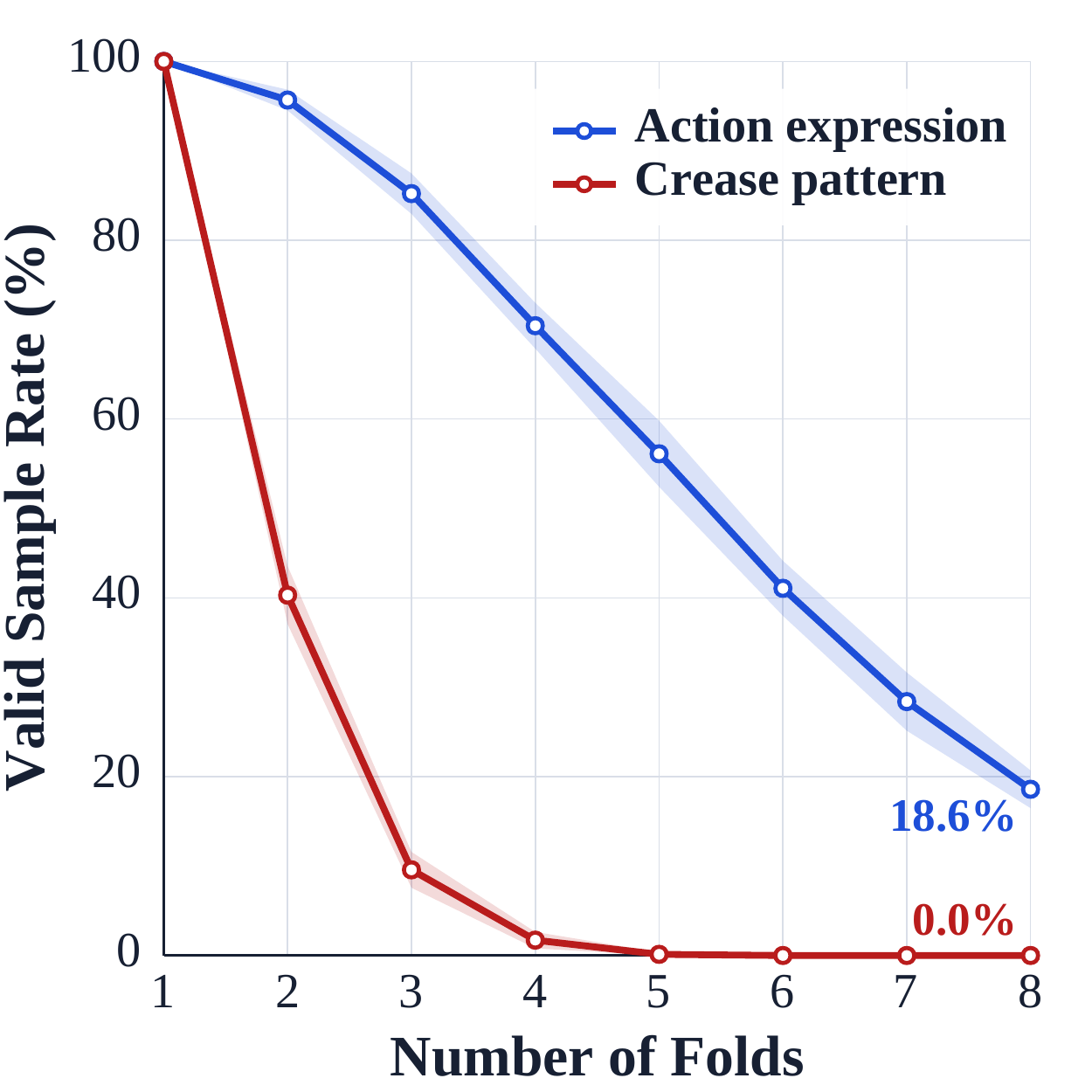}
    \caption{\textbf{Validity under increasing compositional complexity.} Valid sample rates of action expressions and crease patterns across 1 to 8 folds, evaluated on 50,000 paired random samples (2,000 per seed across 25 random seeds). Shaded regions show mean ± standard deviation across 25 random seeds. 
    }
\label{fig:random_representation_validity}
\end{wrapfigure}

\BENCH{} is organized as a collection of folding trajectories represented in the grammar introduced in Section~\ref{origami_grammar}.
Each trajectory contains a program expression, its intermediate FOLD states, and the corresponding rendered observations, as shown in Figure~\ref{fig:origamibench}, enabling supervision at both the program and individual state-transition levels.
A key advantage of our action language is that it provides a scalable way to generate physically meaningful crease patterns while abstracting away low-level physical constraints. Rather than specifying crease geometry directly, each action transforms the current paper state, producing creases consistent with its geometry and layer configuration while preserving the intermediate states that lead to the final pattern. As shown in Figure~\ref{fig:random_representation_validity}, randomly sampled statements are substantially more robust to increasing compositional complexity than direct crease-pattern generation, retaining an 18.6\% validity rate at eight folds while the latter approaches zero; full experimental details are provided in Appendix~\ref{app:random_validity}.

Building on this representation, we generate data in three ways: through manual annotation, random program sampling, and LLM-based program generation. Table~\ref{tab:data_statistics} summarizes the number of trajectories and program-length statistics for each subset.

\textbf{Manually annotated data.}
To ground \BENCH{} in realistic, human-designed folding procedures, we construct the manually annotated subset from step-by-step origami instructions.\footnote{These are obtained from free online resources. An example is given in \url{https://www.origamicraze.com/free-ebook/100origami.pdf}.} We build an interface around the \BENCH{} environment (cf. Appendix~\ref{appendix:annotation_tool}) to interactively reproduce each instruction by specifying fold actions on the current paper state and observing the resulting states. For each step, we record the program expression, intermediate crease pattern, and rendered paper state (front and back views), forming an executable folding trajectory.

\textbf{Randomly generated data.} 
To broaden the coverage of the program space and introduce diverse folding trajectories, we generate programs by randomly sampling statements from our grammar. The sampled programs are executed sequentially in our environment workflow, producing both the complete program and its intermediate folding states.

\begin{wraptable}{r}{0.4\textwidth}
\centering
\scriptsize
\setlength{\tabcolsep}{2.5pt}
\renewcommand{\arraystretch}{1.05}
\caption{Composition and program lengths of \BENCH{}. Program lengths are reported as minimum / mean / maximum number of statements across trajectories.}
\label{tab:data_statistics}
\begin{tabular}{@{}lcc@{}}
\toprule
\textbf{Subset} &
\shortstack{\textbf{Trajectories}} &
\shortstack{\textbf{Statements}\\\textbf{per traj.}} \\
\midrule
Manually Annotated & 25  & 4 / 7.7 / 13 \\
Randomly Generated & 80  & 1 / 3.3 / 8  \\
LLM-generated      & 80  & 2 / 4.8 / 8  \\
\midrule
Total              & 185 & 1 / 4.6 / 13 \\
\bottomrule
\end{tabular}
\end{wraptable}

We select 80 valid trajectories from the random representation validity experiment in Figure~\ref{fig:random_representation_validity} for inclusion in \BENCH{}. Further details of the sampling and validation procedure are provided in Appendix~\ref{appendix:random_generation}.

\textbf{LLM-generated data.}
To expand the coverage of recognizable flat-folded shapes, we use manually annotated examples as in-context demonstrations and prompt an LLM to generate new data together with natural-language instructions and corresponding program expressions. The LLM can directly interact with our environment workflow to execute its generated programs and iteratively revise invalid program expressions based on execution feedback. Through this process, we obtain 80 valid trajectories in total. Further details of the prompting and generation procedure are provided in Appendix~\ref{appendix:llm_generation}.

\subsection{Evaluation Tasks}
\begin{figure*}[h]
\vspace{0.5cm}
\begin{center}

\includegraphics[width=\textwidth, trim=0.cm 0.cm 0.cm 4.cm]{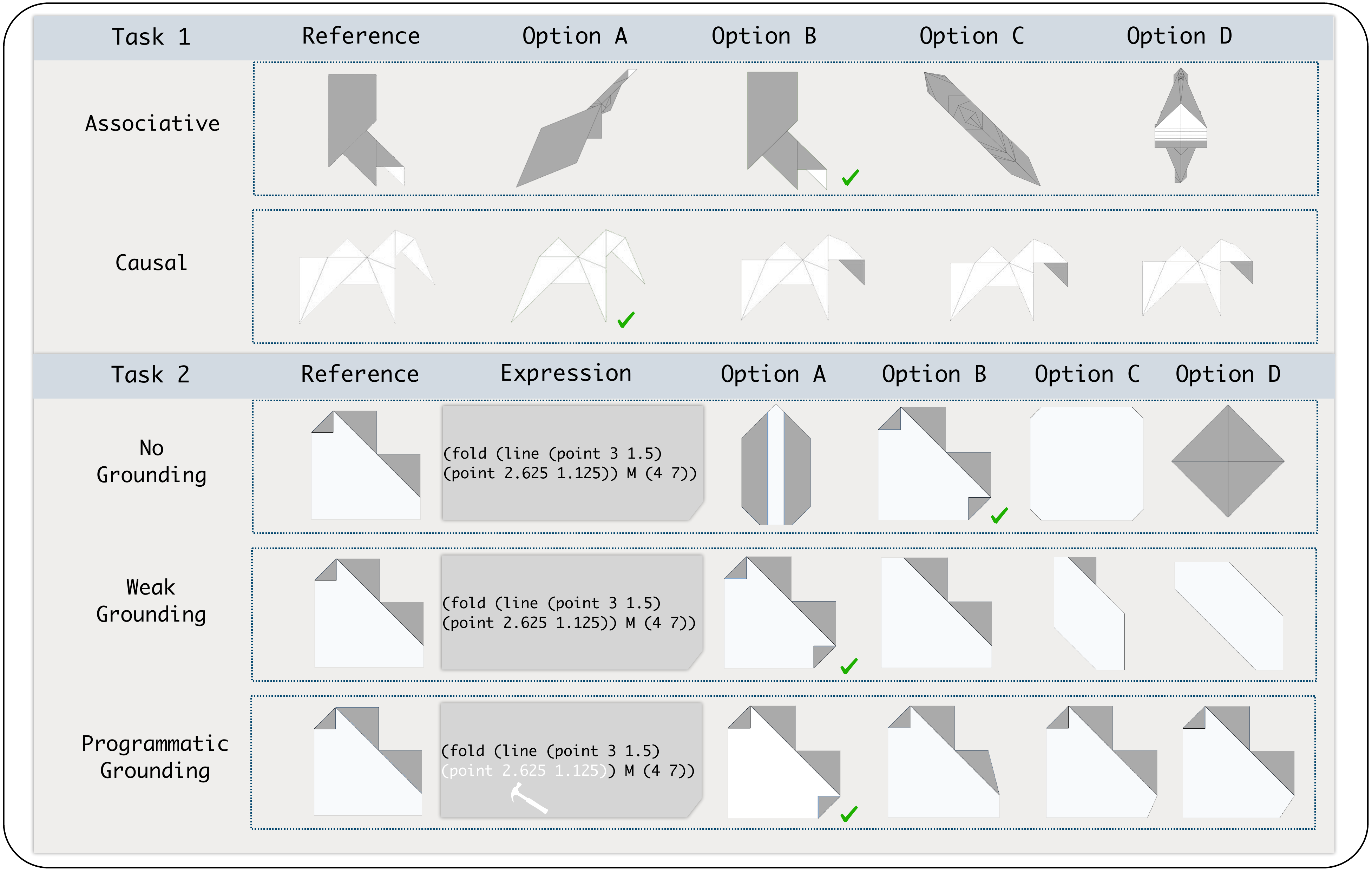}
\end{center}
\caption{\textbf{Construction of evaluation instances.} Each row presents a multiple-choice next-state prediction instance, consisting of a reference origami state and four candidate successors. For programmatic grounding, the intervention is highlighted in white.}
\label{fig:figure_tasks}
\end{figure*}
\BENCH{} evaluates two complementary capabilities: reasoning about local folding transitions and grounding program expressions in their geometric effects. The task variants are designed to progressively remove visual shortcuts and isolate whether models reason about the transition itself and, when available, the semantics of the generating program. Both are formulated as multiple-choice next-state prediction. Given a current rendered state \(S_t\) and candidate successors \(C_t\), the model must identify the ground-truth next state \(S_{t+1}\). The tasks differ in whether the generating statement \(A_t\) is observed.

\noindent\textbf{Task 1: Transition Reasoning.}
The model observes \(S_t\) but not \(A_t\), and must identify the correct successor. We consider two distractor constructions. In \textbf{Task 1a (Associative)}, distractors are sampled from unrelated trajectories. In \textbf{Task 1b (Causal)}, they are sampled from other states of the same trajectory, reducing coarse visual shortcuts and requiring finer reasoning about the local transition. Task 1 is constructed from the Flat-Folder data; dataset and task-construction details are provided in Appendices~\ref{app:dataset} and~\ref{appendix:task1}, respectively.

\noindent\textbf{Task 2: Program Grounding.}
The model additionally observes \(A_t\) and must identify the state resulting from its application to \(S_t\). We evaluate three levels of grounding. In \textbf{Task 2a (No Grounding)}, distractors come from unrelated trajectories, so the task can naively be solved through associative visual matching without relying on the expression. In \textbf{Task 2b (Weak Grounding)}, distractors come from the same trajectory, increasing the value of the symbolic action. In \textbf{Task 2c (Programmatic Grounding)}, distractors are generated by intervening on \(A_t\) and executing the modified expressions from the same state \(S_t\). Interventions modify the abstract syntax tree of a program expression, namely points, layer selections, individual \texttt{fold}s, or the structure of \texttt{combine} expressions. The structured action language therefore enables controlled, semantically meaningful distractors that isolate whether the model is sensitive to the specific geometric effect of the program. The data used for Task 2 are detailed in Appendices~\ref{appendix:annotation_tool}, \ref{appendix:random_generation}, and~\ref{appendix:llm_generation}, respectively, with task-construction details provided in Appendix~\ref{app:onestep}.

{\section{Related Work}\label{sec:related}}
\textbf{Visual and physical reasoning benchmarks.}
Existing benchmarks probe complementary aspects of visual, spatial, algorithmic, and physical reasoning. RAVEN~\citep{zhang2019raven} and ARC-AGI~\citep{chollet2019measure,arc-agi-3} test abstract rule induction, CLEVR compositional reasoning~\citep{johnson2017clevr}, Bongard-LOGO visual concept induction from procedural programs~\citep{nie2020bongard}, and SMART-101 abstraction and spatial reasoning through procedurally generated puzzles~\citep{cherian2023deep}. Physion evaluates physical outcome prediction from videos~\citep{bear2021physion}, while PhyBlock studies physically grounded reasoning and multi-step planning through 3D block assembly~\citep{ma2025phyblock}. Unlike these benchmarks, OrigamiBench evaluates recovery of an executable generative procedure that reconstructs a target object through state-transforming actions.

\textbf{Constructive solid geometry and procedural generation benchmarks.}
Recent work increasingly represents visual content through executable programs.
D$^2$CSG reconstructs CAD shapes as CSG compositions of primitives \citep{yu2023d}, while ArcPro and PyTorchGeoNodes recover procedural shape programs from 3D observations \citep{huang2025arcpro,stekovic2025pytorchgeonodes}.
VLMaterial similarly predicts procedural material graphs from images \citep{li2025vlmaterial}.
In the forward direction, Infinigen and ProcTHOR use procedural generators to synthesize large-scale environments \citep{raistrick2023infinite,deitke2022}.
Procedural reasoning also appears in activity understanding: CaptainCook4D and Ego-Exo4D model multi-step skilled activities and their temporal structure \citep{peddi2024captaincook4d,grauman2024ego}.
These works recover largely declarative descriptions of shape or appearance, use programs for data generation, or recognize observed procedures.
In contrast, OrigamiBench requires recovering an action-oriented sequential program whose execution progressively transforms the origami state, requiring reasoning over both the final geometry and intermediate states.

\textbf{Origami reasoning benchmarks.}
ORIGAMISPACE evaluates MLLMs on pattern prediction, multi-step reasoning, spatial relations, and end-to-end crease-pattern code generation~\citep{xu2025origamispace}, while GamiBench studies 2D-to-3D spatial reasoning through folded-state prediction, viewpoint validation, and impossible-pattern detection~\citep{spencer2025gamibench}.
These benchmarks primarily evaluate reasoning from crease patterns, intermediate folding information, or candidate folded states; even ORIGAMISPACE's generative task targets crease-pattern code.
OrigamiBench~\citep{agarwal2026origamibench} instead evaluates synthesis of the folding procedure itself through an explicit action language in which individual folds and their compositions form executable sequences. 

\textbf{Origami generation systems.}
Computational origami design predates recent learning-based approaches.
Lang introduced algorithmic origami-base design and TreeMaker \citep{lang1996computational}, while Tachi developed methods such as Freeform Origami and Origamizer for generating crease patterns that realize prescribed 3D shapes \citep{tachi2010freeform,tachi2009origamizing}.
Akitaya et al.\ addressed the complementary problem of recovering folding instructions from flat-foldable crease patterns by simplifying the pattern and reversing the resulting transformations \citep{akitaya2013generating}.
More recently, COrigami generates flat-foldable crease patterns from natural-language descriptions using a hybrid geometric and learning-based pipeline \citep{zahavy2026corigami}, while Learn2Fold formulates origami synthesis as conditional program induction with language-model proposals, a learned graph world model, and symbolic feasibility checks \citep{huang2026learn2fold}.
Whereas prior systems address geometric design, crease-pattern generation, or instruction recovery, OrigamiBench systematizes the evaluation of origami synthesis as executable sequential program generation.

{\section{Experimental Results}\label{sec:experiments}}
We conduct two evaluations that both assess the ability of vision-language models (VLMs) to understand and reason about origami state transitions, but use different task formulations. The first evaluates one-step transition reasoning from visual states, contrasting associative matching with causal next-state discrimination. The second additionally provides the generating program expression to assess whether models can ground its symbolic structure in the geometric transformation it induces.

\begin{figure}[t]
\centering
\includegraphics[width=0.7\textwidth, trim=0.cm 0.cm 0.cm 0.cm]{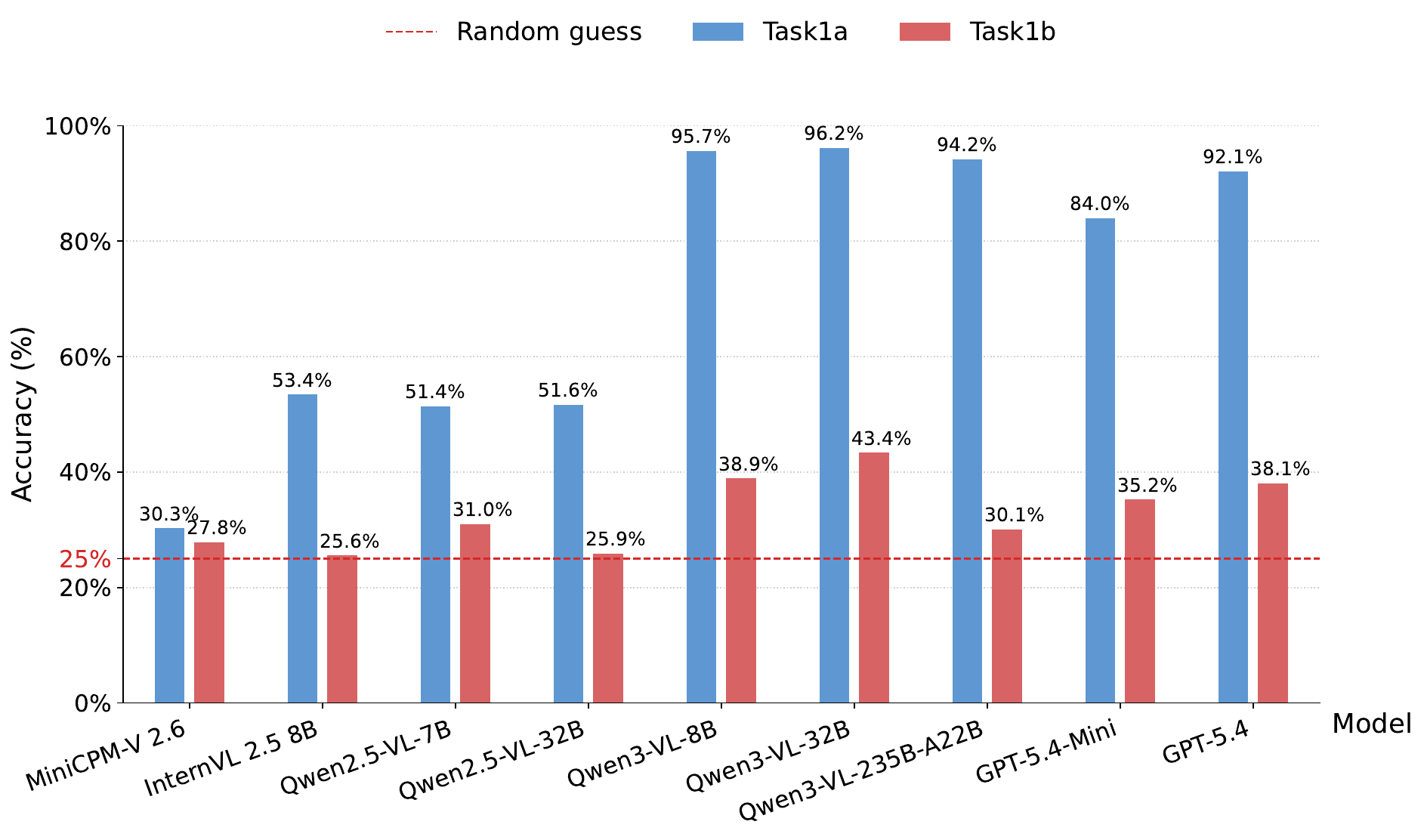}
\caption{Accuracy on Task 1 (Transition Reasoning) across models for the Associative (Task 1a) and Causal (Task 1b) settings. The dashed line marks the 25\% random-guess baseline.}
\label{fig:task1}
\end{figure}

\begin{figure}[t]
\centering
\includegraphics[width=0.72\textwidth, trim=0.cm 0.cm 0.cm 0.cm]{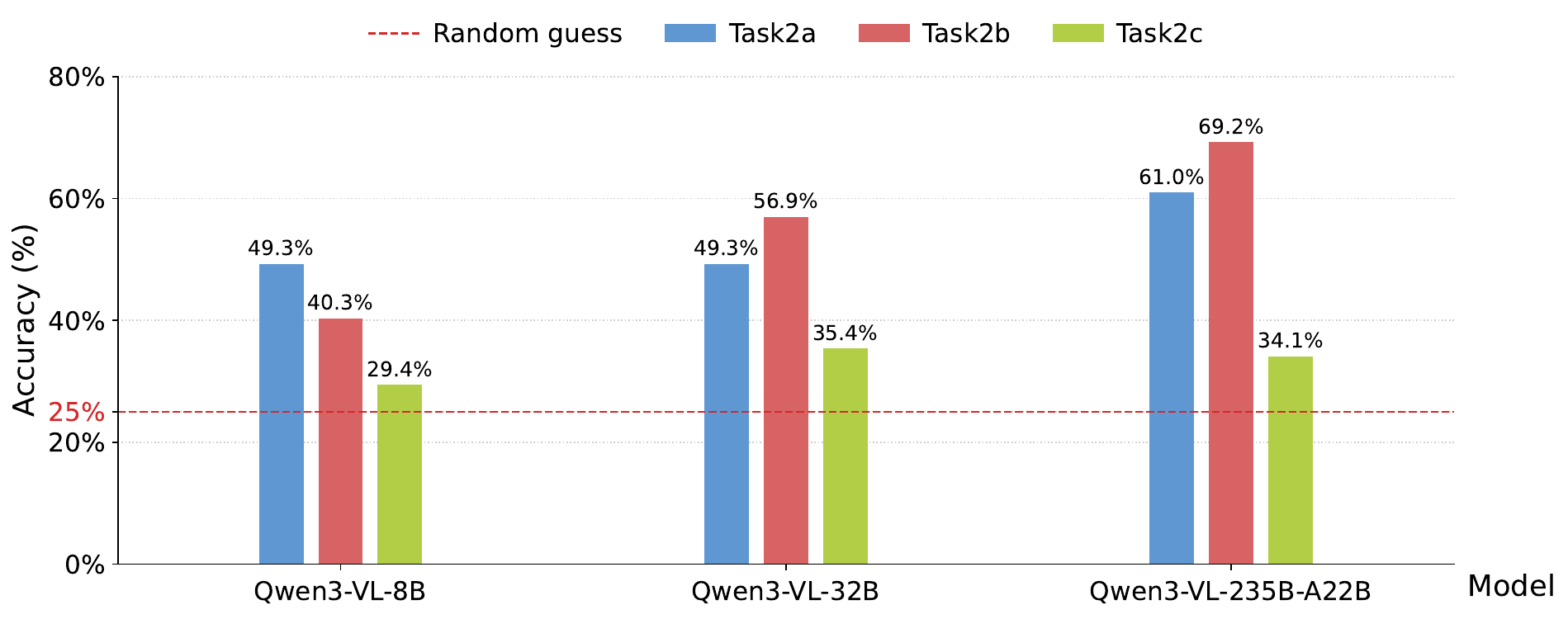}
\caption{Accuracy on Task 2 (Program Grounding) across the No Grounding (Task 2a), Weak Grounding (Task 2b), and Programmatic Grounding (Task 2c) settings. The dashed line marks the corresponding random-guess baselines. Detailed results for individual intervention types in Task 2c are reported in Appendix~\ref{app:task2_perturbations}.}
\label{fig:task2}
\end{figure}

\textbf{Transition reasoning: Associative vs. Causal.}
This evaluation probes whether VLMs can predict the next origami state from a given state without seeing the folding action. However, success may reflect either associative visual matching or genuine causal understanding of the transition between consecutive folding states. By comparing the Associative and Causal variants of Task 1, we disentangle these two behaviors. All trials offer four candidate states, setting the random-guess baseline at 25\%.

On the \textbf{Task 1a (Associative)} setting, where distractors are sampled from unrelated origami trajectories, recent VLMs perform remarkably well.
Qwen3-VL-32B and Qwen3-VL-8B achieve 96.2\% and 95.7\% respectively, followed by Qwen3-VL-235B-A22B at 94.2\% and GPT-5.4 at 92.1\%. GPT-5.4-Mini also reaches 84.0\%. By contrast, InternVL 2.5 8B and the Qwen2.5-VL models obtain around 51\%--53\%, while MiniCPM-V 2.6 reaches 30.0\%, only slightly above the 25\% random baseline.
These results indicate that when candidates come from unrelated folding trajectories, larger VLMs can often identify the correct next state.\footnote{Task 1a benefits from dataset diversity, as visually distinct distractors often allow the correct next state to be identified through visual matching alone.}

The picture changes markedly in \textbf{Task 1b (Causal)}, where all candidates are drawn from the same folding trajectory.
Qwen3-VL-32B achieves the best performance at 43.4\%, followed by Qwen3-VL-8B and GPT-5.4 at roughly 38\%. In comparison, the remaining models fall below 32\%, with several barely exceeding the 25\% random baseline.

The contrast between the two variants is pronounced among the strongest models. For example, Qwen3-VL-32B drops from 96.2\% on Task 1a to 43.4\% on Task 1b, while Qwen3-VL-235B-A22B drops from 94.2\% to 30.1\%. Moreover, the lower Task 1b accuracy of Qwen3-VL-235B-A22B relative to the 8B and 32B variants suggests that model scale alone does not consistently improve transition reasoning.  Additional experiments with eight candidate options are provided in Appendix~\ref{appendix:task1}.

\textbf{Program grounding: Effect of symbolic supervision.}
We next test whether VLMs can use the program expression generating an observed geometric transformation, rather than relying solely on visual correspondences between states. We intervene on the generating action expression $A_t$ at different levels of its abstract syntax tree. This allows us to test whether the symbolic expression provides useful information for identifying the correct transition and, more importantly, whether models can distinguish between different types of symbolic interventions.

Figure~\ref{fig:task2} reports results for the three Qwen3-VL variants (8B, 32B, 235B-A22B) across No Grounding (\textbf{Task 2a}), Weak Grounding (\textbf{Task 2b}), and Programmatic Grounding (\textbf{Task 2c}). Task 2c corresponds to the action intervention setting. All three main tasks use four candidates, giving a 25\% random baseline.

On \textbf{Task 2a (No Grounding)}, where distractors are sampled from unrelated trajectories, the models achieve 49.3\%, 49.3\%, and 61.0\% for Qwen3-VL-8B, Qwen3-VL-32B, and Qwen3-VL-235B-A22B, respectively. These results show that visual transition cues alone can often be sufficient to identify the correct state, although performance remains lower than in the associative setting of Task 1.

Providing the program expression improves performance in \textbf{Task 2b (Weak Grounding)}, despite the fact that the candidates are drawn from the same trajectory and are therefore visually more challenging. Qwen3-VL-32B reaches 56.9\%, while Qwen3-VL-235B-A22B reaches 69.2\%; Qwen3-VL-8B obtains 40.3\%. Thus, for the larger models in particular, access to the program expression provides useful additional information for distinguishing the correct state when visual differences between candidates are less distinctive. This improvement suggests that larger models can use the symbolic expression to detect the presence of a change in the generating action, even when purely visual matching is less reliable. Performance drops significantly in \textbf{Task 2c (Programmatic Grounding)}, where the interventions are varied across the structure of the program expression. Qwen3-VL-8B, Qwen3-VL-32B, and Qwen3-VL-235B-A22B achieve 29.4\%, 35.4\%, and 34.1\% accuracy, providing only a modest improvement over the random-guess baseline. The gap between Task 2b and Task 2c indicates limited understanding of the specific intervention being applied. As in Task 1, scaling alone does not resolve this difficulty: Larger models benefit more from weak grounding, but none achieves strong performance on fine-grained programmatic grounding. Additional Task 2c experiments with individual intervention types are provided in Appendix~\ref{app:onestep}.

{\section{Discussion}\label{sec:future}}
We have introduced a high-level language for describing the synthesis process of flat-foldable origami and, building on it, developed a benchmark to systematically probe the capabilities of modern VLMs. Our current experimental analysis focuses on single-step semantics. Future work will extend the evaluation to multi-step semantics, abstraction, and planning, as well as to orthogonal directions such as 3D origami. More broadly, benchmarks of this kind can help advance causal machine learning, physical AI, and neuro-symbolic programming by providing controlled settings in which models must connect structured representations with the mechanisms governing physical transformations. Such capabilities are relevant to domains including manufacturing, robotics, graphics, and protein folding. 

{\section{Author Contributions}}
N.A. contributed to the development of the environment and annotation tool, data collection, and the writing of the appendices. Y.W. and X.G. contributed to data collection, experimental analysis, and the writing of Section 2.3 and the appendices. A.G. and Y.H. contributed to data collection. V.C. contributed to the formalization of the grammar and the initial drafting of Sections 2 and 4. M.L. contributed to Figures 1 and 5 and to the initial draft of the related work section. Y.X.D. provided computational resources for the experiments. E.S. established, coordinated, and mentored the team throughout the project; secured funding for computational resources; conceptualized the benchmark, the language and the experimental tasks; wrote the abstract and introduction; and contributed to the revisions of the manuscript.

{\section{Acknowledgements}}
This work received funding from the European Research Council (ERC) under the Horizon Europe research and innovation programme (MSCA-GF grant agreement N° 101149800), National Natural Science Foundation of China (U23A20311,62206245), and Jiangsu Science Foundation Leading-edge Technology Program (BK20232003). This work used the Delta and DeltaAI compute resources at the National Center for Supercomputing Applications at the University of Illinois through allocation CIS261499 from the Advanced Cyberinfrastructure Coordination Ecosystem: Services \& Support (ACCESS) program, which is supported by U.S. National Science Foundation grants \#2138259, \#2138286, \#2138307, \#2137603, and \#2138296. The authors would like to thank Robert J. Lang and Raphael Romero for fruitful discussion on the formalization of the language of origami. \nocite{boerner2023access}

\bibliography{ref}
\bibliographystyle{icml2026}


\appendix
{\section{Reproducibility Details for  Figure~\ref{fig:random_representation_validity}}\label{app:random_validity}
}
In Figure~\ref{fig:random_representation_validity}, we generated 50,000 paired random samples using 25 deterministic random seeds, with 2,000 samples per seed. For each sample, the number of folds was drawn uniformly from 1 to 8. For each fold, the two endpoints of the fold line were sampled independently from a $0.001$-spaced grid over $[0,3]\times[0,3]$, and the mountain/valley assignment was sampled with equal probability. Coincident endpoints were resampled, as they do not define a valid fold line. Only atomic folds were considered; \texttt{combine}, library operations, and paper turns were excluded. The complete sequence of fold parameters was sampled before evaluation and shared between the two representations, ensuring that each paired sample used the same fold count, fold lines, and mountain/valley assignments.

For the action-expression representation, the sampled folds were applied sequentially to a blank square, with each action selecting all currently available paper layers. After each action, OrigamiEnv verified that Rabbit Ear could execute the fold, Flat-Folder could produce a flat-foldable state, and the resulting layer order was compatible with the requested mountain/valley assignment. An action expression was considered valid only if all folds were successfully executed and the final folded state was renderable. For the crease-pattern representation, the same sampled fold lines were placed simultaneously on the initial square. A crease pattern was considered valid only if Flat-Folder could solve the complete pattern and OrigamiEnv could render the resulting folded state. The Flat-Folder search was limited to 10,000 states.

No rejection sampling or repair was performed for invalid folds. If the $k$-th action in an action expression failed, execution terminated immediately and the entire expression was recorded as invalid, with $k$ recorded as its failure step. Likewise, a crease pattern was recorded as invalid if its evaluation failed. Invalid samples remained in the denominator and were not replaced; the experiment simply proceeded to the next independently sampled pair. Thus, the reported validity rate measures the probability that a randomly sampled $n$-fold input can be executed successfully under each representation without repairing or resampling invalid folds.

{\section{Data: Flat-Folder}\label{app:dataset}}
We leverage the dataset provided in Flat-Folder dataset, which includes 366 entries, covering a diverse range of origami models. The data are categorized along two dimensions: semantic type and complexity level. Specifically, complexity is determined by the sum of the number of vertices and edges, with thresholds set at 300 and 800 (Easy: 0-300, Medium: 301-800, Hard: $>$800). Table~\ref{tab:dataset_new_structure} presents all classified data, including each model's name, semantic category, and complexity level. Table~\ref{tab:dataset_stats} provides a statistical overview of the dataset's composition, detailing the distribution of the entries across the semantic categories and complexity levels.

\small
\begin{longtable}{l l l}
\caption{Dataset Structure}
\label{tab:dataset_new_structure}\\
\toprule
\textbf{Name} & \textbf{Semantic} & \textbf{Complexity} \\
\midrule
\endfirsthead

\multicolumn{3}{c}{{\tablename\ \thetable{} -- continued from previous page}} \\
\toprule
\textbf{Name} & \textbf{Semantic} & \textbf{Complexity} \\
\midrule
\endhead

\midrule
\multicolumn{3}{r}{{Continued on next page}} \\
\endfoot

\bottomrule
\endlastfoot

010\_chan\_Beaver & animals-aquatic & Easy \\
065\_komatsu\_Dolphin & animals-aquatic & Easy \\
030\_tanakaoripa\_fish\_base & animals-aquatic & Medium \\
129\_komatsu\_Shore\_Crab & animals-aquatic & Medium \\
221\_winston\_Cardinal\_Tetra & animals-aquatic & Medium \\
226\_kamiya\_Hermit\_Crab & animals-aquatic & Medium \\
278\_ku\_Crab\_(base) & animals-aquatic & Medium \\
362\_nakamura\_Tuna & animals-aquatic & Medium \\
020\_imai\_Taiyaki\_Fish & animals-aquatic & Hard \\
037\_imai\_Shrimp\_v2 & animals-aquatic & Hard \\
077\_ku\_Lobster\_1-8b & animals-aquatic & Hard \\
162\_imai\_Carp & animals-aquatic & Hard \\
239\_imai\_Squilla\_Mantis\_Shrimp & animals-aquatic & Hard \\
279\_ku\_Crab\_(shaped) & animals-aquatic & Hard \\
299\_brandon\_Japanese\_Spiny\_Lobster\_(jsl) & animals-aquatic & Hard \\
300\_imai\_Japanese\_Spiny\_Lobster\_v2 & animals-aquatic & Hard \\
313\_maeng\_Shrimp & animals-aquatic & Hard \\
329\_maeng\_Lobster & animals-aquatic & Hard \\
355\_nakamura\_Octopus & animals-aquatic & Hard \\
\midrule

016\_tanakaoripa\_enarc-crane\_final & animals-bird & Easy \\
082\_traditionaloripa\_4\_Birdbase & animals-bird & Easy \\
096\_komatsu\_Horned\_Owl & animals-bird & Easy \\
158\_katsuta\_Sparrow & animals-bird & Easy \\
163\_kei\_Party\_Parrot & animals-bird & Easy \\
236\_kamiya\_Crane & animals-bird & Easy \\
244\_komatsuoripa\_Smallbird\_Base & animals-bird & Easy \\
003\_ku\_Triangle\_Bird & animals-bird & Medium \\
089\_traditionaloripa\_9\_Birdbase & animals-bird & Medium \\
136\_katsuta\_Owl & animals-bird & Medium \\
159\_kamiya\_Little\_Bird & animals-bird & Medium \\
160\_komatsu\_Little\_Bird & animals-bird & Medium \\
173\_imai\_Canabn & animals-bird & Hard \\
331\_gyosh\_Condor\_v1 & animals-bird & Hard \\
\midrule

018\_komatsu\_Butterfly\_Base & animals-insects & Easy \\
281\_kamiya\_Long-Headed\_Locust & animals-insects & Easy \\
064\_tanakaoripa\_Butterfly\_Base & animals-insects & Medium \\
147\_kamiya\_Butterfly\_NS & animals-insects & Medium \\
261\_lang\_Post\_Generic\_Beetle\_Opus\_696 & animals-insects & Medium \\
339\_haruka\_Beetle & animals-insects & Medium \\
029\_imai\_Swallowtail\_Butterfly\_v2 & animals-insects & Hard \\
046\_imai\_Prosopocoilus\_Inclinatus\_v3-1 & animals-insects & Hard \\
053\_imai\_Leaf\_Insect\_v2 & animals-insects & Hard \\
085\_imai\_Flying\_Grasshopper & animals-insects & Hard \\
095\_imai\_Mantis\_v2-8 & animals-insects & Hard \\
104\_imai\_Metallifer & animals-insects & Hard \\
108\_ryan\_Titan\_Beetle & animals-insects & Hard \\
112\_imai\_Cicada\_v1 & animals-insects & Hard \\
119\_kamiya\_Wasp & animals-insects & Hard \\
120\_imai\_Forest\_Scorpion & animals-insects & Hard \\
134\_table\_Scorpion & animals-insects & Hard \\
142\_ku\_Butterfly & animals-insects & Hard \\
149\_imai\_Chilean\_Stag\_Beetle & animals-insects & Hard \\
175\_shuki\_Dragonfly & animals-insects & Hard \\
195\_kamiya\_Butterfly\_TH & animals-insects & Hard \\
198\_imai\_Cetoniinae & animals-insects & Hard \\
214\_bodo\_Ant & animals-insects & Hard \\
219\_imai\_Copris\_Ochus & animals-insects & Hard \\
234\_bodo\_Dragonfly & animals-insects & Hard \\
252\_imai\_Chalcosoma\_Chiron & animals-insects & Hard \\
256\_bodo\_Ladybug & animals-insects & Hard \\
260\_imai\_Lucanus\_Maculifemoratus & animals-insects & Hard \\
273\_kamiya\_Cyclomantis\_Metalifer & animals-insects & Hard \\
283\_imai\_Dragonfly & animals-insects & Hard \\
285\_maeng\_Cyclommatus\_Elaphus & animals-insects & Hard \\
291\_imai\_Chalcosoma\_Caucasus & animals-insects & Hard \\
306\_maeng\_Weta & animals-insects & Hard \\
310\_imai\_Dorcus\_hopei\_binodulosus & animals-insects & Hard \\
314\_minh\_Butterfly & animals-insects & Hard \\
321\_maeng\_Dicranocephalus\_wallichii & animals-insects & Hard \\
333\_imai\_Violin\_Beetle & animals-insects & Hard \\
336\_maeng\_Mantis & animals-insects & Hard \\
343\_maeng\_Orchid\_Mantis & animals-insects & Hard \\
350\_maeng\_Sitophilus\_oryzae\_v2 & animals-insects & Hard \\
\midrule

045\_traditionaloripa\_Pig & animals-mammals & Easy \\
048\_komatsu\_Dutch\_Rabbit & animals-mammals & Easy \\
055\_komatsu\_Sheep & animals-mammals & Easy \\
113\_komatsu\_Fox & animals-mammals & Easy \\
124\_xiao\_Horse & animals-mammals & Easy \\
135\_xiao\_Lazy\_Tiger & animals-mammals & Easy \\
138\_komatsu\_Cat & animals-mammals & Easy \\
141\_kei\_kitsune & animals-mammals & Easy \\
144\_xiao\_Bull & animals-mammals & Easy \\
183\_komatsu\_Panda & animals-mammals & Easy \\
207\_komatsu\_Horse & animals-mammals & Easy \\
212\_shuki\_Bear\_Cub & animals-mammals & Easy \\
218\_kei\_Tanuki & animals-mammals & Easy \\
227\_komatsu\_Rabbit & animals-mammals & Easy \\
259\_kei\_Father\_Cat & animals-mammals & Easy \\
327\_origamivan\_Alpaca\_v1 & animals-mammals & Easy \\
338\_origamivan\_Mouse & animals-mammals & Easy \\
359\_fung\_I\_Heart\_Cat\_v2\_(base) & animals-mammals & Easy \\
005\_ku\_Kitten & animals-mammals & Medium \\
011\_ku\_Bull & animals-mammals & Medium \\
034\_fung\_Rabbit\_Head\_v4 & animals-mammals & Medium \\
081\_komatsu\_Lion & animals-mammals & Medium \\
105\_komatsu\_Rhino & animals-mammals & Medium \\
146\_katsuta\_Lazy\_Cat & animals-mammals & Medium \\
148\_komatsu\_Wolf & animals-mammals & Medium \\
171\_komatsu\_Macaque & animals-mammals & Medium \\
196\_komatsu\_Giraffe & animals-mammals & Medium \\
211\_luibobby\_Rabbit & animals-mammals & Medium \\
217\_komatsu\_Hippopotamus & animals-mammals & Medium \\
225\_xiao\_Chipmunk & animals-mammals & Medium \\
230\_fung\_Rabbit\_Head\_v3 & animals-mammals & Medium \\
237\_komatsu\_Tiger & animals-mammals & Medium \\
243\_luibobby\_Giraffe & animals-mammals & Medium \\
248\_komatsu\_Dog & animals-mammals & Medium \\
271\_xiao\_Raccoon & animals-mammals & Medium \\
337\_ku\_Rabbit & animals-mammals & Medium \\
347\_morisawa\_Raccoon & animals-mammals & Medium \\
353\_fukuroi\_Dalmatian & animals-mammals & Medium \\
360\_fung\_I\_Heart\_Cat\_v2\_(shaped) & animals-mammals & Medium \\
123\_shuki\_Bison & animals-mammals & Hard \\
143\_shuki\_Asian\_Elephant & animals-mammals & Hard \\
154\_shuki\_Camel & animals-mammals & Hard \\
204\_bodo\_Tiger & animals-mammals & Hard \\
\midrule

004\_traditional\_Crane & animals-others & Easy \\
021\_chan\_Simple\_Dragon & animals-others & Easy \\
070\_ku\_Pygmy\_Jerboa & animals-others & Easy \\
073\_komatsu\_Squirrel & animals-others & Easy \\
083\_kei\_Cheems\_Meme & animals-others & Easy \\
235\_xiao\_Tiny\_Dragon & animals-others & Easy \\
040\_komatsu\_Gentle\_Dragon & animals-others & Medium \\
042\_furutaoripa\_katatsumuri\_(snail) & animals-others & Medium \\
101\_ku\_Turkey\_2-1 & animals-others & Medium \\
137\_kamiya\_Deinonychus & animals-others & Medium \\
180\_kamiya\_Pheonix\_v2-0 & animals-others & Medium \\
289\_kamiya\_Kentrosaurus & animals-others & Medium \\
311\_ku\_HJ\_Dragon & animals-others & Medium \\
351\_ku\_Three\_Headed\_Dragon & animals-others & Medium \\
050\_ku\_Ryu\_Zin\_Jr & animals-others & Hard \\
057\_brandon\_Dragon\_HP & animals-others & Hard \\
059\_bodo\_Leopard\_Frog & animals-others & Hard \\
068\_bodo\_Sarcosuchus & animals-others & Hard \\
132\_shuki\_Giganotosaurus\_v4 & animals-others & Hard \\
139\_imai\_Red\_Headed\_Centipede & animals-others & Hard \\
140\_imai\_Red\_Headed\_Centipede\_40 & animals-others & Hard \\
181\_kamiya\_Pheonix\_v3-0 & animals-others & Hard \\
182\_kamiya\_Pheonix\_v3-5 & animals-others & Hard \\
200\_shuki\_Zoanoid\_Dragon\_v1 & animals-others & Hard \\
201\_shuki\_Zoanoid\_Dragon\_v2 & animals-others & Hard \\
209\_imai\_Whip\_Spider & animals-others & Hard \\
229\_imai\_Scorpion\_v2 & animals-others & Hard \\
250\_juan\_Velociraptor & animals-others & Hard \\
268\_imai\_Tarantula & animals-others & Hard \\
277\_maeng\_Camel\_Spider & animals-others & Hard \\
332\_arisawa\_Dragon\_2017 & animals-others & Hard \\
366\_imai\_Vinegaroon\_v2 & animals-others & Hard \\
\midrule

344\_ku\_Lizard & animals-reptiles & Medium \\
066\_brandon\_Rattlesnake\_HP & animals-reptiles & Hard \\
240\_juan\_Galapagos\_Tortoise & animals-reptiles & Hard \\
265\_kamiya\_Loggerhead\_Sea\_Turtle & animals-reptiles & Hard \\
266\_kamiya\_Turtle & animals-reptiles & Hard \\
342\_kei\_Rain\_Frog & animals-reptiles & Hard \\
\midrule

071\_imai\_Sailor\_Suit & clothing & Medium \\
\midrule

013\_mitanioripa\_Medal & decorations & Easy \\
130\_fung\_Christmas\_Tree\_v2 & decorations & Easy \\
028\_chan\_The\_Universal\_Symbol & decorations & Medium \\
199\_fung\_Classic\_Snowman\_v1 & decorations & Medium \\
131\_ku\_Christmas\_Tree & decorations & Hard \\
161\_tanaka\_star & decorations & Hard \\
\midrule

009\_ku\_Unassigned\_Triangle\_Pleat & geometry & Easy \\
015\_ku\_Iso-Area\_Half\_Square & geometry & Easy \\
043\_ku\_Quad\_Overlap & geometry & Easy \\
054\_tanakaoripa\_Hexagon & geometry & Easy \\
076\_kei\_Fibonacci\_Vortex & geometry & Easy \\
091\_ku\_4\_Stripes\_1 & geometry & Easy \\
092\_ku\_4\_Stripes\_2 & geometry & Easy \\
093\_ku\_4\_Stripes\_3 & geometry & Easy \\
098\_RSorigami\_Inside\_Out & geometry & Easy \\
125\_brown\_seamless4x4 & geometry & Easy \\
292\_tanaka\_Aperiodic\_Monotile\_1 & geometry & Easy \\
293\_bodo\_Aperiodic\_Monotile\_1 & geometry & Easy \\
294\_ku\_Aperiodic\_Monotile\_1 & geometry & Easy \\
295\_tanaka\_Aperiodic\_Monotile\_2 & geometry & Easy \\
296\_ku\_Aperiodic\_Monotile\_2 & geometry & Easy \\
297\_tanaka\_Aperiodic\_Monotile\_3 & geometry & Easy \\
325\_bernhayes\_Unassigned\_45°\_Crossover & geometry & Easy \\
049\_lang\_Egg13-3\_Tessellation & geometry & Medium \\
086\_miwu\_5x5\_Flippable\_Pixel\_Grid\_1 & geometry & Medium \\
087\_miwu\_5x5\_Flippable\_Pixel\_Grid\_2 & geometry & Medium \\
094\_ku\_4\_Stripes\_4 & geometry & Medium \\
164\_ku\_Pleats\_Exponential & geometry & Medium \\
191\_fish\_Arrowhead\_Tessellation & geometry & Medium \\
361\_lang\_Squaring\_the\_Circle & geometry & Medium \\
027\_ku\_8x8\_Flippable\_Pixel\_Grid & geometry & Hard \\
031\_boice\_Cat's\_Eye\_Tessellation & geometry & Hard \\
033\_lang\_Hyperbolic\_Limit & geometry & Hard \\
072\_tanakaoripa\_Infinity\_2 & geometry & Hard \\
145\_beber\_Hexagonal\_Tessellation\_of\_Dodecagons\_\#1 & geometry & Hard \\
157\_beber\_Square\_Tessellation\_of\_Dodecagons\_\#3 & geometry & Hard \\
167\_fish\_Desert\_Tessellation & geometry & Hard \\
169\_beber\_Hexagonal\_Tessellation\_of\_Dodecagons\_\#2 & geometry & Hard \\
179\_beber\_Rotated\_3-4-6-4 & geometry & Hard \\
184\_tanaka\_notitle & geometry & Hard \\
192\_beber\_Dodecagon\^{}2 & geometry & Hard \\
202\_beber\_Hexagonal\_Tessellation\_of\_Dodecagons\_\#3 & geometry & Hard \\
213\_beber\_fractal32 & geometry & Hard \\
222\_beber\_Penrose\_Triangle\_\#2 & geometry & Hard \\
231\_beber\_Dodecagon\^{}3 & geometry & Hard \\
242\_beber\_om & geometry & Hard \\
254\_beber\_Menger\_Sponge\_Level\_2\_I & geometry & Hard \\
255\_beber\_Menger\_Sponge\_Level\_2\_II & geometry & Hard \\
262\_beber\_Square\_Tessellation\_of\_Dodecagons\_\#2 & geometry & Hard \\
269\_beber\_Penrose+ & geometry & Hard \\
276\_beber\_Dodecagon\^{}3-4 & geometry & Hard \\
284\_beber\_Sierpinski-Penrose\_Triangle\_\#2 & geometry & Hard \\
364\_luibobby\_Cross\_with\_hydrangea\_tessellation & geometry & Hard \\
\midrule

197\_kei\_Pray & gestures & Easy \\
290\_kei\_Hand & gestures & Medium \\
\midrule

002\_traditional\_Kabuto & item & Easy \\
052\_traditionaloripa\_House & item & Easy \\
067\_traditionaloripa\_damashi\_bune & item & Easy \\
187\_fung\_Heart\_in\_a\_House\_(Flat)\_v1 & item & Easy \\
341\_jared\_Bitcoin & item & Easy \\
017\_jorge\_Tato\_Coaster & item & Medium \\
106\_kei\_7\_Segment\_Display\_(off) & item & Medium \\
107\_kei\_7\_Segment\_Display\_(on) & item & Medium \\
330\_ku\_F16 & item & Medium \\
346\_origamivan\_BattleTank & item & Medium \\
349\_kei\_Umbrella\_2022 & item & Medium \\
177\_chan\_Mortal\_Kombat\_Emblem & item & Hard \\
232\_luibobby\_MLG\_Glasses & item & Hard \\
298\_kei\_TIE\_Interceptor\_(Star\_Wars) & item & Hard \\
307\_kei\_Arc\_Trooper\_(Star\_Wars) & item & Hard \\
\midrule

012\_ku\_4x4\_Grid\_Unassigned & others & Easy \\
026\_furutaoripa\_orihazuru & others & Easy \\
035\_tachioripa\_kamehameha & others & Easy \\
038\_traditionaloripa\_Yakko & others & Easy \\
056\_tachioripa\_Hoodman & others & Easy \\
062\_furutaoripa\_houou & others & Easy \\
069\_kei\_Disappearing\_Contrails & others & Easy \\
074\_miuraoripa\_Miura-ori & others & Easy \\
088\_komatsu\_Papillion & others & Easy \\
117\_tanaka\_Dice\_15 & others & Easy \\
122\_kei\_jitome & others & Easy \\
156\_xiao\_Makima & others & Easy \\
174\_kei\_Steps & others & Easy \\
186\_kei\_Sus & others & Easy \\
223\_luibobby\_Discord & others & Easy \\
282\_kei\_Circus & others & Easy \\
301\_boxhard\_Assigned\_Crossover & others & Easy \\
302\_boxhard\_Assigned\_Clause & others & Easy \\
303\_boxhard\_Assigned\_Split & others & Easy \\
316\_boxhard\_Unassigned\_Clause & others & Easy \\
317\_boxhard\_Unassigned\_Crossover & others & Easy \\
318\_boxhard\_Unassigned\_Split & others & Easy \\
348\_nakamura\_Cheat\_Dice & others & Easy \\
041\_lang\_5-Fold\_2-Layer\_Weave & others & Medium \\
047\_tanakaoripa\_Infinity & others & Medium \\
060\_ku\_1-Piece\_Star\_Atarbus & others & Medium \\
061\_kei\_koishi & others & Medium \\
115\_tanaka\_Dice\_13 & others & Medium \\
116\_tanaka\_Dice\_8 & others & Medium \\
133\_txst\_himeno & others & Medium \\
205\_xiao\_Morgan & others & Medium \\
228\_kei\_seigaiha & others & Medium \\
253\_luibobby\_Reddit & others & Medium \\
267\_kei\_Plat & others & Medium \\
272\_juhobodo\_Poser & others & Medium \\
324\_stebleton\_Hammersley's\_See\_Saw & others & Medium \\
328\_kei\_Useless\_Resistance & others & Medium \\
358\_ku\_13\_Stripes & others & Medium \\
019\_lang\_Rings8 & others & Hard \\
022\_boice\_Sword\_Plate\_Armor\_40 & others & Hard \\
024\_komatsu\_Girl-bp-01 & others & Hard \\
025\_lang\_Golden\_Weave & others & Hard \\
039\_boice\_Ara\_Ara & others & Hard \\
063\_imai\_Japanese\_Drone\_v2 & others & Hard \\
100\_kamiya\_Ryujin\_2-1 & others & Hard \\
172\_tanaka\_irt & others & Hard \\
188\_boicehanke\_Square\_Twist\_Samurai & others & Hard \\
233\_tanaka\_equalwidth & others & Hard \\
270\_ku\_HJ\_Rex & others & Hard \\
275\_imai\_Sushi & others & Hard \\
304\_boxhard\_Assigned\_Solvable & others & Hard \\
308\_dequin\_Igris & others & Hard \\
319\_boxhard\_Unassigned\_Solvable & others & Hard \\
354\_lang\_rings4 & others & Hard \\
\midrule

121\_komatsuoripa\_rafflesia & plants & Easy \\
078\_imai\_Maple\_Leaf\_v1 & plants & Medium \\
079\_imai\_Maple\_Leaf\_v2 & plants & Medium \\
080\_imai\_Maple\_Leaf\_v3 & plants & Medium \\
287\_ku\_Four\_Leaf\_Clover & plants & Medium \\
334\_jared\_Green\_Leaf & plants & Medium \\
\midrule

103\_koh\_Footballer & roles & Easy \\
114\_kei\_Kirby & roles & Easy \\
118\_katsuta\_Fox\_Wedding\_Cub & roles & Easy \\
128\_kamiya\_Sleipnir & roles & Easy \\
216\_kamiya\_Minotaur & roles & Easy \\
315\_kei\_Marnie\_(Pokemon) & roles & Easy \\
014\_ku\_HJ\_Girl & roles & Medium \\
090\_kei\_Man & roles & Medium \\
099\_katsuta\_Fox\_Wedding\_Bride & roles & Medium \\
109\_yty\_Pochita & roles & Medium \\
110\_katsuta\_Fox\_Wedding\_Groom & roles & Medium \\
111\_kamiya\_Wizard & roles & Medium \\
127\_katsuta\_Unicorn & roles & Medium \\
150\_kei\_Sailor\_Moon & roles & Medium \\
168\_xiao\_Mermaid & roles & Medium \\
170\_kamiya\_Bahamut & roles & Medium \\
194\_xiao\_DaJi & roles & Medium \\
206\_kamiya\_Divine\_Boar & roles & Medium \\
210\_fung\_Angel\_v1 & roles & Medium \\
215\_xiao\_Miku\_2021 & roles & Medium \\
220\_fung\_Portrait\_of\_an\_Origamist\_v1 & roles & Medium \\
224\_bodo\_Buff\_doge & roles & Medium \\
246\_xiao\_Surfing\_Girl & roles & Medium \\
249\_kei\_Poyoyo & roles & Medium \\
258\_kamiya\_Cait\_Sith & roles & Medium \\
263\_ku\_Angel & roles & Medium \\
264\_xiao\_YuMeiRen & roles & Medium \\
280\_xiao\_Lying\_Girl & roles & Medium \\
288\_xiao\_Skating\_Girl & roles & Medium \\
312\_kamiya\_Kzinssie\_Type\_2 & roles & Medium \\
320\_kamiya\_Unicorn & roles & Medium \\
323\_chan\_M.O.D.O.K. & roles & Medium \\
356\_kei\_Dodoco\_(Genshin\_Impact) & roles & Medium \\
365\_ku\_Nazgul\_v8.1 & roles & Medium \\
032\_komatsuoripa\_Minotaur & roles & Hard \\
044\_kei\_Angel & roles & Hard \\
075\_bodo\_Dwarf & roles & Hard \\
126\_yuchao\_Bing\_Dwen\_Dwen & roles & Hard \\
155\_fish\_Man\_Bat & roles & Hard \\
165\_shuki\_Guyver\_Unit\_III & roles & Hard \\
166\_shuki\_Guyver\_Unit\_I & roles & Hard \\
176\_fish\_Hephaetus & roles & Hard \\
185\_imai\_Jack\_Skellington & roles & Hard \\
189\_shuki\_Evangelion\_Unit\_01\_2009 & roles & Hard \\
190\_shuki\_Evangelion\_Unit\_01\_2013 & roles & Hard \\
193\_chan\_Spider\_Man & roles & Hard \\
203\_hanke\_Archangel\_Michael & roles & Hard \\
208\_kei\_General\_Grievous & roles & Hard \\
238\_kei\_Danbo & roles & Hard \\
241\_fung\_Amabie\_v2 & roles & Hard \\
245\_bodo\_Batman\_on\_his\_Bike & roles & Hard \\
247\_kamiya\_Tenma\_H-7 & roles & Hard \\
274\_kei\_Keqing\_(Genshin\_Impact) & roles & Hard \\
286\_carrotkei\_Keied\_Keqing & roles & Hard \\
305\_kamiya\_Winged\_Kirin & roles & Hard \\
309\_xiaokei\_Keied\_Raiden\_Shogun & roles & Hard \\
322\_kei\_Cerberus\_(Helltaker) & roles & Hard \\
326\_imai\_Flying\_Mantis & roles & Hard \\
335\_kei\_Hatsune\_Miku & roles & Hard \\
340\_imai\_new\_Flying\_Mantis & roles & Hard \\
345\_haruka\_Ayanami\_Rei\_3.0 & roles & Hard \\
352\_haruka\_Venusaur & roles & Hard \\
363\_kei\_Hatoba\_Tsugu\_(Vtuber) & roles & Hard \\
\midrule

006\_ku\_Thirds\_Pinwheel & toys & Easy \\
007\_ku\_Pinwheel\_Pockets & toys & Easy \\
008\_ku\_Robot & toys & Easy \\
058\_ku\_Color-change\_Pinwheel\_1 & toys & Easy \\
084\_ku\_4x4\_Checkerboard & toys & Easy \\
102\_traditionaloripa\_Pinwheel & toys & Easy \\
151\_ku\_Iso-Area\_Throwing\_Star\_1 & toys & Easy \\
153\_ku\_Iso-Area\_Throwing\_Star\_3 & toys & Easy \\
178\_xiao\_Teddy\_Bear & toys & Easy \\
257\_xiao\_Fairy\_Doll & toys & Easy \\
152\_ku\_Iso-Area\_Throwing\_Star\_2 & toys & Medium \\
251\_fung\_Kokeshi\_Doll\_v4 & toys & Medium \\
036\_ku\_Checkerboard\_26x26 & toys & Hard \\
357\_chan\_8x8\_Checkerboard & toys & Hard \\
\midrule

001\_traditional\_Sailboat & vehicles & Easy \\
023\_itagaki\_Motorcycle & vehicles & Hard \\
\midrule

051\_kei\_CP\_PLZ & words & Easy \\
097\_kei\_CP\_THX & words & Medium \\

\end{longtable}

\begin{table}[t]
\caption{OrigamiBench Dataset Statistics by Semantic Category and Complexity Level}
\label{tab:dataset_stats}
\centering
\small
\begin{tabular}{l c c c c}
\toprule
\textbf{Semantic Category} & \textbf{Easy} & \textbf{Medium} & \textbf{Hard} & \textbf{TOTAL}\\
\midrule
animal-aquatic & 2 & 6 & 11 & 19 \\
animal-bird & 7 & 5 & 2 & 14 \\
animal-insects & 2 & 4 & 34 & 40 \\
animal-mammals & 18 & 21 & 4 & 43 \\
animal-reptiles & 0 & 1 & 5 & 6 \\
animal-others & 6 & 8 & 18 & 32 \\
clothing & 0 & 1 & 0 & 1 \\
decorations & 2 & 2 & 2 & 6 \\
geometry & 17 & 7 & 23 & 47 \\
gestures & 1 & 1 & 0 & 2 \\
item & 5 & 6 & 4 & 15 \\
plants & 1 & 5 & 0 & 6 \\
roles & 6 & 28 & 29 & 63\\
toys & 10 & 2 & 2 & 14 \\
vehicles & 1 & 0 & 1 & 2 \\
words & 1 & 1 & 0 & 2 \\
others & 23 & 15 & 16 & 54 \\
\midrule
\textbf{TOTAL} & 102 & 113 & 151 & 366 \\
\bottomrule
\end{tabular}
\end{table}

{\section{Data: Manual Annotation}
\label{appendix:annotation_tool}}
We developed an interactive annotation tool to collect executable folding programs for origami. The interface mirrors the physical folding process, allowing annotators to follow standard origami tutorials while recording each fold in the same order in which it would be performed on paper. This ensures that the resulting annotations directly represent the folding trajectory and are compatible with the program representation used in our benchmark.

\subsection{Annotation Workflow}

For each folding step, annotators follow three stages: \textbf{Set Up Fold}, \textbf{Review and Edit}, and \textbf{Validate and Build Sequence}. They may additionally use \textbf{Quick Actions} during annotation and \textbf{Finalize and Save} once the origami is complete.

In \textbf{Set Up Fold}, annotators specify the fold visually on the current origami state. They can select the fold type (e.g., mountain or valley), apply predefined library folds and bases (library folds applied to the starting square of origami), draw the fold line on an interactive grid, and select the paper layers involved in the operation.

The resulting symbolic expression is shown in \textbf{Review and Edit}, where annotators can inspect and manually refine its parameters when necessary. In \textbf{Validate and Build Sequence}, the tool executes the fold using the same renderer employed by the benchmark and displays the resulting state. Invalid operations are rejected with an error message, allowing annotators to correct the fold before adding it to the trajectory.

\textbf{Quick Actions} support operations such as undoing previous folds and flipping the paper, making it possible to follow tutorials that alternate between the front and back of the origami without restarting the annotation.

\begin{figure}[htbp]
\centering
\vspace{30pt}
\includegraphics[width=\textwidth, trim=0.cm 2.cm 0.cm 2.cm]{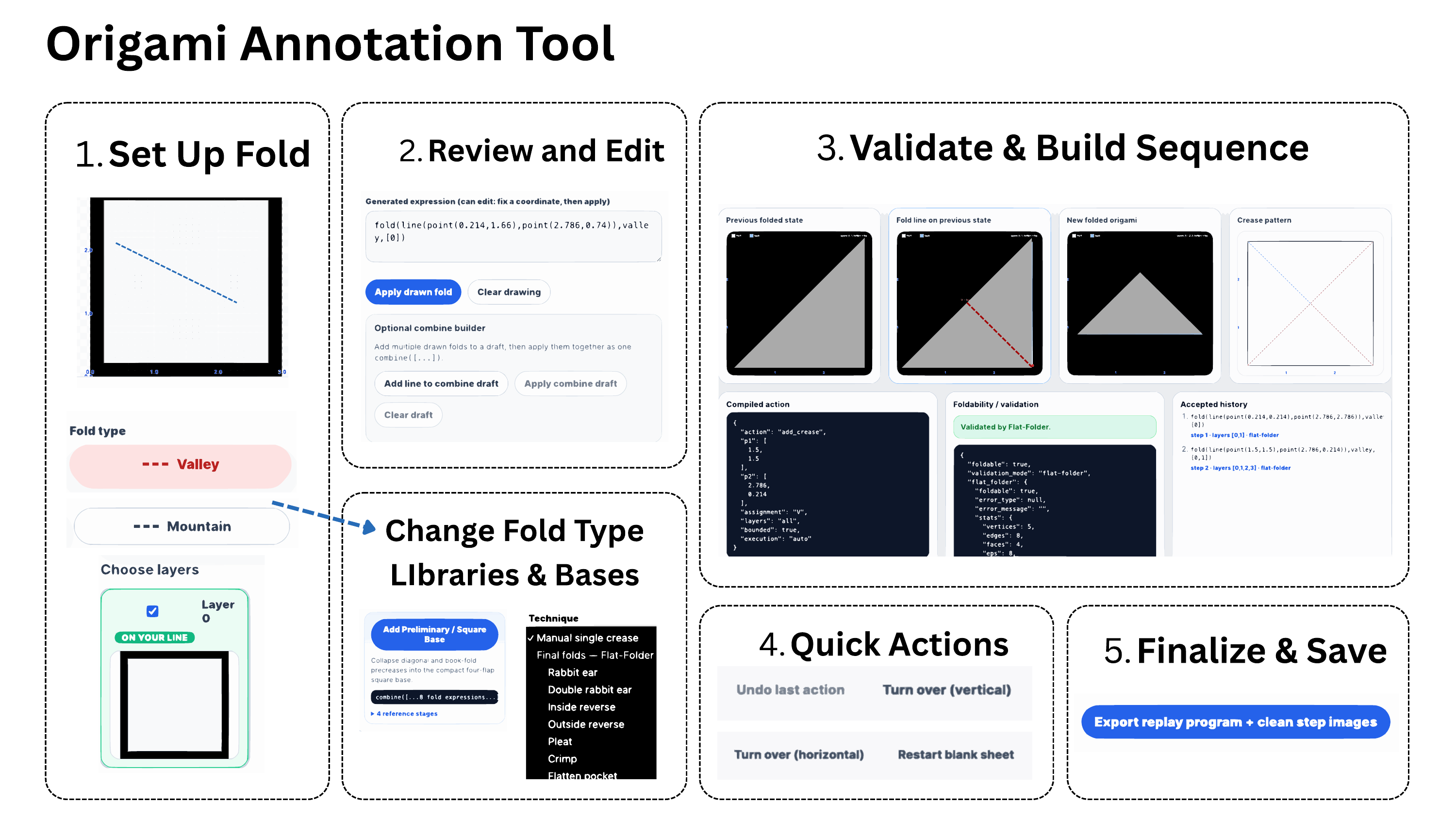}
\vspace{3pt}
\caption{Annotation tool workflow.}
\label{fig:annotation_tool}
\end{figure}

After completing the origami, the annotator saves the full program together with the intermediate visual states. For every step, the tool stores the program expression, front and back renderings, and crease pattern in PNG and SVG formats. Consequently, each annotation provides both the complete executable folding trajectory and its corresponding visual states. Figure~\ref{fig:annotation_tool} summarizes the workflow.

{\section{Data: Random Generation}
\label{appendix:random_generation}}

The randomly generated subset of \BENCH{} is constructed from the valid action expression experiment (details in Appendix~\ref{app:random_validity}). As described, each program is sampled in full and then executed sequentially from a blank sheet, without resampling or repairing invalid actions. For programs that execute successfully, we record the complete program, intermediate folding states, and corresponding rendered observations as an executable folding trajectory. We select 80 such valid trajectories for inclusion in \BENCH{}.

{\section{Data: LLM Generation}
\label{appendix:llm_generation}}
\begin{longtable}{l l}
\caption{Current dataset structure (all programs). Source indicates source family.}
\label{tab:dataset_current_structure}\\
\toprule
\textbf{Name} & \textbf{Source} \\
\midrule
\endfirsthead

\multicolumn{2}{c}{{\tablename\ \thetable{} -- continued from previous page}} \\
\toprule
\textbf{Name} & \textbf{Source} \\
\midrule
\endhead

\midrule
\multicolumn{2}{r}{{Continued on next page}} \\
\endfoot

\bottomrule
\endlastfoot

ma\_001\_bear\_cub & manually-annotated \\
ma\_002\_bear\_face & manually-annotated \\
ma\_003\_dogonkennel & manually-annotated \\
ma\_004\_heart & manually-annotated \\
ma\_005\_bear\_face\_2 & manually-annotated \\
ma\_006\_easy\_heart & manually-annotated \\
ma\_007\_panda & manually-annotated \\
ma\_008\_pencil & manually-annotated \\
ma\_009\_apple & manually-annotated \\
ma\_010\_bananas & manually-annotated \\
ma\_011\_fortune\_teller & manually-annotated \\
ma\_012\_chick & manually-annotated \\
ma\_013\_dog & manually-annotated \\
ma\_014\_duck\_1 & manually-annotated \\
ma\_015\_duck\_2 & manually-annotated \\
ma\_016\_fox & manually-annotated \\
ma\_017\_rabbit & manually-annotated \\
ma\_018\_rabbit\_2 & manually-annotated \\
ma\_019\_topical\_fish\_2 & manually-annotated \\
ma\_020\_talking\_dog & manually-annotated \\
ma\_021\_envelope & manually-annotated \\
ma\_022\_sailboat & manually-annotated \\
ma\_023\_cup & manually-annotated \\
ma\_024\_fortune\_teller\_2 & manually-annotated \\
ma\_025\_pengiun & manually-annotated \\
\midrule

synthetic\_001\_grammar\_00001 & random-atomic-grammar \\
synthetic\_002\_grammar\_00004 & random-atomic-grammar \\
synthetic\_003\_grammar\_00005 & random-atomic-grammar \\
synthetic\_004\_grammar\_00007 & random-atomic-grammar \\
synthetic\_005\_grammar\_00008 & random-atomic-grammar \\
synthetic\_006\_grammar\_00009 & random-atomic-grammar \\
synthetic\_007\_grammar\_00012 & random-atomic-grammar \\
synthetic\_008\_grammar\_00019 & random-atomic-grammar \\
synthetic\_009\_grammar\_00022 & random-atomic-grammar \\
synthetic\_010\_grammar\_00023 & random-atomic-grammar \\
synthetic\_011\_grammar\_00025 & random-atomic-grammar \\
synthetic\_012\_grammar\_00031 & random-atomic-grammar \\
synthetic\_013\_grammar\_00035 & random-atomic-grammar \\
synthetic\_014\_grammar\_00039 & random-atomic-grammar \\
synthetic\_015\_grammar\_00040 & random-atomic-grammar \\
synthetic\_016\_grammar\_00058 & random-atomic-grammar \\
synthetic\_017\_grammar\_00060 & random-atomic-grammar \\
synthetic\_018\_grammar\_00064 & random-atomic-grammar \\
synthetic\_019\_grammar\_00067 & random-atomic-grammar \\
synthetic\_020\_grammar\_00072 & random-atomic-grammar \\
synthetic\_021\_grammar\_00073 & random-atomic-grammar \\
synthetic\_022\_grammar\_00074 & random-atomic-grammar \\
synthetic\_023\_grammar\_00075 & random-atomic-grammar \\
synthetic\_024\_grammar\_00077 & random-atomic-grammar \\
synthetic\_025\_grammar\_00079 & random-atomic-grammar \\
synthetic\_026\_grammar\_00084 & random-atomic-grammar \\
synthetic\_027\_grammar\_00085 & random-atomic-grammar \\
synthetic\_028\_grammar\_00091 & random-atomic-grammar \\
synthetic\_029\_grammar\_00094 & random-atomic-grammar \\
synthetic\_030\_grammar\_00095 & random-atomic-grammar \\
synthetic\_031\_grammar\_00097 & random-atomic-grammar \\
synthetic\_032\_grammar\_00101 & random-atomic-grammar \\
synthetic\_033\_grammar\_00111 & random-atomic-grammar \\
synthetic\_034\_grammar\_00112 & random-atomic-grammar \\
synthetic\_035\_grammar\_00115 & random-atomic-grammar \\
synthetic\_036\_grammar\_00118 & random-atomic-grammar \\
synthetic\_037\_grammar\_00128 & random-atomic-grammar \\
synthetic\_038\_grammar\_00129 & random-atomic-grammar \\
synthetic\_039\_grammar\_00133 & random-atomic-grammar \\
synthetic\_040\_grammar\_00138 & random-atomic-grammar \\
synthetic\_041\_grammar\_00139 & random-atomic-grammar \\
synthetic\_042\_grammar\_00140 & random-atomic-grammar \\
synthetic\_043\_grammar\_00147 & random-atomic-grammar \\
synthetic\_044\_grammar\_00153 & random-atomic-grammar \\
synthetic\_045\_grammar\_00158 & random-atomic-grammar \\
synthetic\_046\_grammar\_00159 & random-atomic-grammar \\
synthetic\_047\_grammar\_00163 & random-atomic-grammar \\
synthetic\_048\_grammar\_00164 & random-atomic-grammar \\
synthetic\_049\_grammar\_00173 & random-atomic-grammar \\
synthetic\_050\_grammar\_00174 & random-atomic-grammar \\
synthetic\_051\_random\_combine\_001 & random-combine \\
synthetic\_052\_random\_combine\_002 & random-combine \\
synthetic\_053\_random\_combine\_003 & random-combine \\
synthetic\_054\_random\_combine\_004 & random-combine \\
synthetic\_055\_random\_combine\_005 & random-combine \\
synthetic\_056\_random\_combine\_006 & random-combine \\
synthetic\_057\_random\_combine\_007 & random-combine \\
synthetic\_058\_random\_combine\_008 & random-combine \\
synthetic\_059\_random\_combine\_009 & random-combine \\
synthetic\_060\_random\_combine\_010 & random-combine \\
synthetic\_061\_random\_combine\_011 & random-combine \\
synthetic\_062\_random\_combine\_012 & random-combine \\
synthetic\_063\_random\_combine\_013 & random-combine \\
synthetic\_064\_random\_combine\_014 & random-combine \\
synthetic\_065\_random\_combine\_015 & random-combine \\
synthetic\_066\_random\_library\_pleat\_001 & random-library \\
synthetic\_067\_random\_library\_pleat\_002 & random-library \\
synthetic\_068\_random\_library\_pleat\_003 & random-library \\
synthetic\_069\_random\_library\_pleat\_004 & random-library \\
synthetic\_070\_random\_library\_pleat\_005 & random-library \\
synthetic\_071\_random\_library\_pleat\_006 & random-library \\
synthetic\_072\_random\_library\_pleat\_007 & random-library \\
synthetic\_073\_random\_library\_pleat\_008 & random-library \\
synthetic\_074\_random\_library\_crimp\_001 & random-library \\
synthetic\_075\_random\_library\_crimp\_002 & random-library \\
synthetic\_076\_random\_library\_crimp\_003 & random-library \\
synthetic\_077\_random\_library\_crimp\_004 & random-library \\
synthetic\_078\_random\_library\_crimp\_005 & random-library \\
synthetic\_079\_random\_library\_crimp\_006 & random-library \\
synthetic\_080\_random\_library\_crimp\_007 & random-library \\
\midrule

ai\_001\_square\_envelope & llm-generated \\
ai\_002\_simple\_house & llm-generated \\
ai\_003\_paper\_cup & llm-generated \\
ai\_004\_compact\_napkin & llm-generated \\
ai\_005\_gate\_panel & llm-generated \\
ai\_006\_letter\_packet & llm-generated \\
ai\_007\_quarter\_tile & llm-generated \\
ai\_008\_long\_bookmark & llm-generated \\
ai\_009\_wide\_belt & llm-generated \\
ai\_010\_diagonal\_ribbon & llm-generated \\
ai\_011\_shield & llm-generated \\
ai\_012\_cut\_gem & llm-generated \\
ai\_013\_coffin & llm-generated \\
ai\_014\_luggage\_tag & llm-generated \\
ai\_015\_ticket\_stub & llm-generated \\
ai\_016\_lantern & llm-generated \\
ai\_017\_rounded\_tablet & llm-generated \\
ai\_018\_acorn & llm-generated \\
ai\_019\_octagon\_badge & llm-generated \\
ai\_020\_pennant & llm-generated \\
ai\_021\_tombstone & llm-generated \\
ai\_022\_rocket\_plaque & llm-generated \\
ai\_023\_tree\_plaque & llm-generated \\
ai\_024\_bell & llm-generated \\
ai\_025\_mushroom\_cap & llm-generated \\
ai\_026\_trapezoid\_bag & llm-generated \\
ai\_027\_boat\_hull & llm-generated \\
ai\_028\_party\_hat & llm-generated \\
ai\_029\_sailboat & llm-generated \\
ai\_030\_mountain\_peak & llm-generated \\
ai\_031\_fox\_head & llm-generated \\
ai\_032\_cat\_head & llm-generated \\
ai\_033\_fish & llm-generated \\
ai\_034\_bird & llm-generated \\
ai\_035\_leaf & llm-generated \\
ai\_036\_single\_sail & llm-generated \\
ai\_037\_triangle\_hat & llm-generated \\
ai\_038\_open\_envelope & llm-generated \\
ai\_039\_flower\_bud & llm-generated \\
ai\_040\_pinwheel\_tile & llm-generated \\
ai\_041\_photo\_frame & llm-generated \\
ai\_042\_arrow\_tile & llm-generated \\
ai\_043\_kite\_badge & llm-generated \\
ai\_044\_candy\_wrapper & llm-generated \\
ai\_045\_diamond\_pocket & llm-generated \\
ai\_046\_rocket & llm-generated \\
ai\_047\_pencil & llm-generated \\
ai\_048\_necktie & llm-generated \\
ai\_049\_surfboard & llm-generated \\
ai\_050\_octagonal\_window & llm-generated \\
ai\_051\_framed\_shield & llm-generated \\
ai\_052\_striped\_gem & llm-generated \\
ai\_053\_capped\_coffin & llm-generated \\
ai\_054\_double\_lantern & llm-generated \\
ai\_055\_ribbon\_medal & llm-generated \\
ai\_056\_arrow\_plaque & llm-generated \\
ai\_057\_faceted\_vase & llm-generated \\
ai\_058\_bottle\_label & llm-generated \\
ai\_059\_temple\_tablet & llm-generated \\
ai\_060\_asymmetric\_crystal & llm-generated \\
ai\_061\_pointed\_tower & llm-generated \\
ai\_062\_small\_bottle & llm-generated \\
ai\_063\_wide\_vase & llm-generated \\
ai\_064\_candle\_holder & llm-generated \\
ai\_065\_crayon & llm-generated \\
ai\_066\_tie\_banner & llm-generated \\
ai\_067\_narrow\_surfboard & llm-generated \\
ai\_068\_key\_tag & llm-generated \\
ai\_069\_column\_capital & llm-generated \\
ai\_070\_arch\_window & llm-generated \\
ai\_071\_layered\_envelope & llm-generated \\
ai\_072\_woven\_square & llm-generated \\
ai\_073\_blossom\_frame & llm-generated \\
ai\_074\_windmill\_badge & llm-generated \\
ai\_075\_diamond\_medal & llm-generated \\
ai\_076\_fox\_mask & llm-generated \\
ai\_077\_fishtail & llm-generated \\
ai\_078\_bird\_wing & llm-generated \\
ai\_079\_leaf\_spear & llm-generated \\
ai\_080\_triangle\_mosaic & llm-generated \\
\end{longtable}

{\section{Task 1}
\label{appendix:task1}}
\subsection{Task Generation}
In Task 1, we construct the multiple-choice tasks using all 366 entries from Flat-Folder data. For each model, we first generate the corresponding one-step crease pattern sequence using the Creasy tool.\footnote{\url{https://github.com/xkevio/Creasy}} The crease patterns are then rendered in our environment to obtain the corresponding folded results, which serve as the ground-truth outputs.

For \textbf{Task 1a (Associative)}, we construct one evaluation instance for each of the 366 origami. The ground-truth successor is paired with distractors randomly sampled from folded states of other origami trajectories, resulting in 366 instances in total. For \textbf{Task 1b (Causal)}, distractors are instead sampled from other states within the same origami trajectory. Since constructing four candidate choices requires sufficiently long trajectories, we restrict Task 1b to the 255 origami whose folding sequences contain more than four states, resulting in 255 instances.

For both variants, each instance consists of the current folded state as input and four candidate successor states (A--D), including one ground-truth successor and three distractors. We balance the position of the correct answer across the four choices to avoid answer-position bias.

\subsection{Prompt}

The one-step task is formulated as a multiple-choice question in which, given a ``Reference’’ origami (current state), the model must select the candidate produced by exactly one additional fold, while the remaining candidates serve as randomly sampled distractors.

\begin{tcolorbox}[breakable, title=Prompt for Associative Task]

You are an intelligent agent capable of understanding 2D geometry, spatial relationships, topological transformations, and origami crease mechanics.
Your task is to sequentially analyze a 2D origami folding process and logically deduce the correct immediate next state from a set of options. \\

IMPORTANT: Your output is parsed by a strict regex parser.
If your format is incorrect, the answer will be rejected. \\

================================ \\
TASK OVERVIEW \\
================================ \\

You are provided with a sequence of five images. The first four are candidate options (A, B, C, D), and the final image is the Reference state.
For every image, the left side displays the FRONT view of the origami, and the right side displays the BACK view. \\

Your objective is to identify which of the four options represents the exact state obtained from the Reference by performing EXACTLY ONE forward fold operation. \\

================================ \\
HOW TO APPROACH THE TASK \\
================================ \\

\begin{itemize}
    \item Analyze the Reference State: Examine the visible outer-layer creases, straight edges, raw paper boundaries, and flaps on BOTH the front and back views. Build a mental 3D model of the current layer ordering.
    
    \item Identify the Discrepancies: Compare each option (A, B, C, D) directly against the Reference state. Identify exactly what parts of the Reference state are identically preserved and what parts have changed (e.g., a flap has moved, new faces are exposed, edges are realigned). Think about the geometry of the shape, not superficial visual similarity.
    
    \item Filter out "Non-Immediate" Steps (Rank by Dependency \& Timeline):
    \begin{itemize}
        \item Reject "Unfolds" (Backward steps): Does the option lack a crease present in the Reference, or show a flap unfolded compared to the Reference? If it represents a step backward in time, eliminate it.
        
        \item Reject "Skips" (Multiple steps): Does the option show multiple distinct flaps moving independently, or a complex structural change spanning disconnected regions? The correct candidate must be strictly EXACTLY ONE forward fold. If it requires two or more sequential independent folds, eliminate it.
    \end{itemize}
    
    \item Reverse Engineer the Operation: For the remaining viable options, determine the exact physical action required (e.g., mountain fold, valley fold, squash fold, reverse fold). Ask yourself:
    \begin{itemize}
        \item Where does the fold originate? Is there a pre-crease (dashed/dotted line) in the Reference state that serves as the hinge?
        \item What specific polygonal region must be isolated and moved?
        \item Does this single fold leave the rest of the origami structure completely unchanged?
    \end{itemize}
    
    \item Verify Physical Foldability (Rank by Irreversibility):
    \begin{itemize}
        \item Layer Legality: Does the action respect current layer ordering? Will this fold force solid paper to clip through itself?
        \item Dual-View Consistency (Front/Back): A fold on the outer layer (front) strictly implies a corresponding geometric change on the reverse side (back). Eliminate options where the front and back geometries contradict each other or violate physical paper volume.
    \end{itemize}
    
    \item Mental Simulation: Mentally simulate the fold from the Reference. What layers move? What rotates? Does it land EXACTLY on your chosen option?
\end{itemize}

================================ \\
IMAGE LEGEND \\
================================ \\

Images are provided in a fixed order:

\begin{enumerate}
    \item Option A (Front \& Back)
    \item Option B (Front \& Back)
    \item Option C (Front \& Back)
    \item Option D (Front \& Back)
    \item Reference State (Front \& Back)
\end{enumerate}

================================ \\
STRICT RULES \\
================================ \\

\begin{itemize}
    \item You MUST evaluate both the FRONT and BACK views simultaneously. A perfectly matching front with an inconsistent back is a wrong answer.
    \item You MUST correctly identify and reject backward steps and skipped steps.
    \item Do NOT choose purely by "most visually similar". A visually identical silhouette might harbor illegal layer-swaps, while a correct single fold can significantly alter the silhouette.
    \item Do NOT output your reasoning text. ONLY output the final format.
\end{itemize}

================================ \\
OUTPUT CONTRACT (CRITICAL) \\
================================ \\

Your final output MUST strictly consist of EXACTLY ONE line containing the format below and NOTHING ELSE:

\[
\boxed{X}
\]

where X is your chosen option: A, B, C, or D. Do not append extra text or punctuation.

\end{tcolorbox}

\begin{tcolorbox}[breakable, title=Prompt for Causal Task]

You are an intelligent agent capable of understanding 2D geometry, spatial relationships, topological transformations, and origami crease mechanics.
Your task is to sequentially analyze a 2D origami folding process and logically deduce the correct immediate next state from a set of options. \\

IMPORTANT: Your output is parsed by a strict regex parser.
If your format is incorrect, the answer will be rejected. \\

================================ \\
TASK OVERVIEW \\
================================ \\

You are provided with a sequence of five images. The first four are candidate options (A, B, C, D), and the final image is the Reference state.
For every image, the left side displays the FRONT view of the origami, and the right side displays the BACK view. \\
All images come from the exact same continuous folding sequence. \\

Your objective is to identify which of the four options represents the exact state obtained from the Reference by performing EXACTLY ONE forward fold operation. \\

================================ \\
HOW TO APPROACH THE TASK \\
================================ \\

\begin{itemize}
    \item Analyze the Reference State: Examine the visible outer-layer creases, straight edges, raw paper boundaries, and flaps on BOTH the front and back views. Build a mental 3D model of the current layer ordering.
    
    \item Identify the Discrepancies: Compare each option (A, B, C, D) directly against the Reference state. Identify exactly what parts of the Reference state are identically preserved and what parts have changed (e.g., a flap has moved, new faces are exposed, edges are realigned). Think about the geometry of the shape, not superficial visual similarity.
    
    \item Filter out "Non-Immediate" Steps (Rank by Dependency \& Timeline):
    \begin{itemize}
        \item Reject "Unfolds" (Backward steps): Does the option lack a crease present in the Reference, or show a flap unfolded compared to the Reference? If it represents a step backward in time, eliminate it.
        
        \item Reject "Skips" (Multiple steps): Does the option show multiple distinct flaps moving independently, or a complex structural change spanning disconnected regions? The correct candidate must be strictly EXACTLY ONE forward fold. If it requires two or more sequential independent folds, eliminate it.
    \end{itemize}
    
    \item Reverse Engineer the Operation: For the remaining viable options, determine the exact physical action required (e.g., mountain fold, valley fold, squash fold, reverse fold). Ask yourself:
    \begin{itemize}
        \item Where does the fold originate? Is there a pre-crease (dashed/dotted line) in the Reference state that serves as the hinge?
        \item What specific polygonal region must be isolated and moved?
        \item Does this single fold leave the rest of the origami structure completely unchanged?
    \end{itemize}
    
    \item Verify Physical Foldability (Rank by Irreversibility):
    \begin{itemize}
        \item Layer Legality: Does the action respect current layer ordering? Will this fold force solid paper to clip through itself?
        \item Dual-View Consistency (Front/Back): A fold on the outer layer (front) strictly implies a corresponding geometric change on the reverse side (back). Eliminate options where the front and back geometries contradict each other or violate physical paper volume.
    \end{itemize}
    
    \item Mental Simulation: Mentally simulate the fold from the Reference. What layers move? What rotates? Does it land EXACTLY on your chosen option?
\end{itemize}

================================ \\
IMAGE LEGEND \\
================================ \\

Images are provided in a fixed order:

\begin{enumerate}
    \item Option A (Front \& Back)
    \item Option B (Front \& Back)
    \item Option C (Front \& Back)
    \item Option D (Front \& Back)
    \item Reference State (Front \& Back)
\end{enumerate}

================================ \\
STRICT RULES \\
================================ \\

\begin{itemize}
    \item You MUST evaluate both the FRONT and BACK views simultaneously. A perfectly matching front with an inconsistent back is a wrong answer.
    \item You MUST correctly identify and reject backward steps and skipped steps.
    \item Do NOT choose purely by "most visually similar". A visually identical silhouette might harbor illegal layer-swaps, while a correct single fold can significantly alter the silhouette.
    \item Do NOT output your reasoning text. ONLY output the final format.
\end{itemize}

================================ \\
OUTPUT CONTRACT (CRITICAL) \\
================================ \\

Your final output MUST strictly consist of EXACTLY ONE line containing the format below and NOTHING ELSE:

\[
\boxed{X}
\]

where X is your chosen option: A, B, C, or D. Do not append extra text or punctuation.

\end{tcolorbox}

\subsection{Qualitative Results} \label{app:one_step_res}

\begin{figure}[t]
\centering
\vspace{2.5cm}

\includegraphics[width=\textwidth, trim=0.cm 2.cm 0.cm 6.cm]{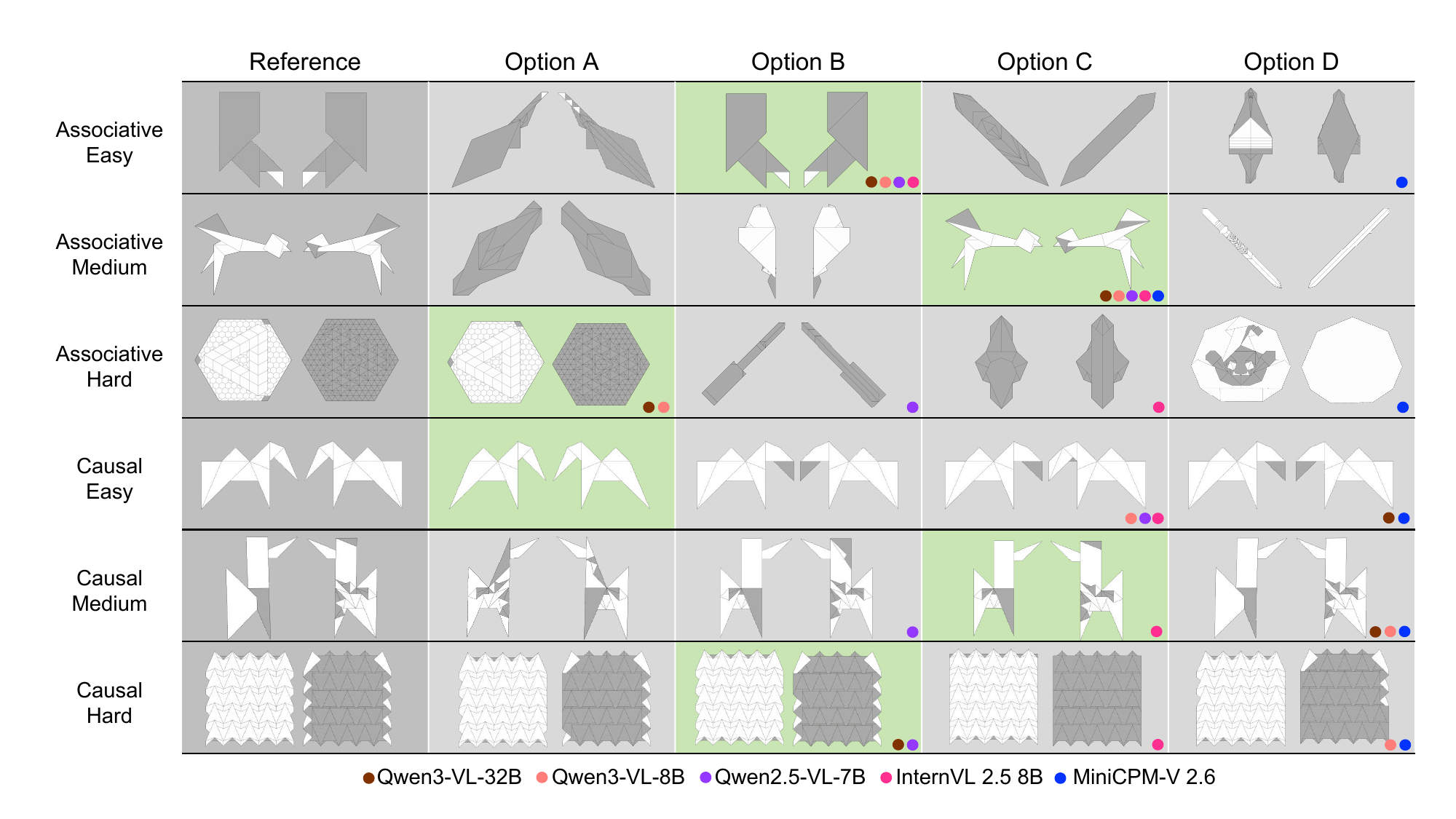}
\vspace{0.2cm}
\caption{Examples for the one-step task. The top and the bottom three rows show Associative and Causal task examples, respectively. Additionally, three task examples with varying degree of complexity are shown for both Associative and Causal. Each task requires to choose the next folding step of the reference origami (first column) from four candidate options (A--D). The correct answer is highlighted using the green background. Coloured circles at the bottom-right corner of each option indicates the prediction made by each model.}
\label{fig:one_step_example}
\end{figure}

Some examples are shown in Figure~\ref{fig:one_step_example}.

\subsection{Additional Experiments} \label{app:one_step_exp}
To further increase the difficulty of the Task 1, we expand the number of candidate options from 4 to 8. For the \textbf{Task 1a (Associative)}, we construct one instance for each of the 366 origami, resulting in 366 evaluation instances, as in the 4-option setting. For the \textbf{Task 1b (Causal)}, the distractors must be sampled from other states within the same origami trajectory. Since constructing eight candidate choices requires sufficiently long trajectories, we restrict this setting to the 243 origami whose folding sequences contain more than eight states. We therefore construct 243 causal evaluation instances. The accuracy of different VLMs under this 8-option setting is summarized in Table~\ref{tab:one_step_results_8_options}.

\begin{table}[t]
\centering
\caption{Accuracy performance of different VLMs on the one-step multiple-choice evaluation task(8 options).}
\vspace{0.2cm}
\begin{tabular}{lcc}
\toprule
\textbf{Model} & Associative & Causal \\
\midrule
Random Guessing & 12.50\% & 12.50\%\\    
\midrule
GPT-5.4 & 89.06\% & 22.63\% \\
GPT-5.4-Mini & 77.86\% & 25.10\% \\
Qwen3-VL-32B & 93.89\% & 32.92\% \\
Qwen3-VL-8B & 91.09\% & 27.16\% \\
Qwen2.5-VL-32B & 46.06\% & 23.05\% \\
Qwen2.5-VL-7B & 70.99\% & 15.64\% \\
InternVL 2.5 8B & 35.62\% & 13.17\% \\
MiniCPM-V 2.6 & 16.28\% & 15.23\%  \\
\bottomrule
\end{tabular}
\label{tab:one_step_results_8_options}
\vspace{-0.4cm}
\end{table}

{\section{Task 2}\label{app:onestep}}
\subsection{Task Generation}
\label{sec:task2-construction}

Task 2 instances are constructed from valid state transitions
\[
    (S_{k-1}, A_k) \longrightarrow S_k,
\]
where \(S_{k-1}\) is the input folded state, \(A_k\) is the corresponding program expression, and \(S_k\) is the ground-truth successor state.

Both input and candidate states are represented by paired front- and back-view renderings.

We use transitions with \(k \geq 4\), excluding very early folding states that are often visually simple. Among the valid trajectories, 68 contain at least four operations. These trajectories provide 173 eligible source transitions.

\paragraph{Task 2a: Cross-sequence distractors.}
For each source transition, the correct option is \(S_k\), while three distractors are sampled from states in three different origami trajectories. To avoid trivial distractors, we sample only sufficiently late states from those trajectories. This produces 173 four-option instances.

\paragraph{Task 2b: Within-sequence distractors.}
Task 2b uses the same 173 source transitions, but samples all three distractors from other states within the same origami trajectory. We exclude the ground-truth successor and avoid using the input state whenever possible. This also produces 173 four-option instances.

\paragraph{Task 2c: Program-generated distractors.}
Instead of retrieving distractors from existing states, Task 2c constructs folding expressions by intervening on $A_k$. Each intervened expression is executed from the exact source state $S_{k-1}$ using the origami environment, and its rendered result is used as a distractor. We divide Task 2c into four variants.

\begin{itemize}
    \item \textbf{Point.} We modify only the two points defining the fold line while preserving the fold type and layer selection. Candidate lines are generated by changing individual coordinates, replacing one endpoint, or translating the original line. Degenerate lines and lines geometrically equivalent to the original fold line are removed. Each retained instance contains the correct result and three distractors with point intervention.

    \item \textbf{Type.} We preserve the fold line and selected layers but reverse the fold type, i.e., mountain is changed to valley or vice versa. Since the fold type has only two possible values, each instance contains two options: the correct result and the result of the reversed fold. An instance is omitted if the reversed operation is invalid or produces no distinct state. Among the 165 retained instances, 72 correct operations are mountain folds ($43.64\%$) and 93 are valley folds ($56.36\%$).

    \item \textbf{Layer.} We preserve the fold line and fold type but modify the set of affected layer identifiers. Candidate layer sets are produced by adding, removing, or replacing layers and by sampling subsets of the layers available in $S_{k-1}$. Each instance contains the correct result and one distractor with layer intervention. Because changing the layer set frequently produces an invalid fold or an almost unchanged rendering, we additionally require that the paper body changes by more than $0.5\%$ of the pixels in at least one of the two views. Rendering annotations are excluded when computing this difference.

    \item \textbf{Mix.} We randomly modify a non-empty subset of the point, type, and layer fields. A distractor may modify a single field or multiple fields simultaneously. Each retained instance contains the correct result and three generated distractors.
\end{itemize}

All generated distractors must be accepted by the origami execution environment. We reject candidates that are invalid, duplicate another option, reproduce the input state, reproduce the ground-truth next state, or match the result of replaying the original expression. Option positions are shuffled using a fixed random seed to make the construction reproducible.

\begin{table}[t]
    \centering
    \small
    \begin{tabular}{llrrrr}
        \toprule
        Variant & Distractor source
        & Distractors & Options & Retained & Skipped \\
        \midrule
        Task 2a & Other origami sequences
        & 3 & 4 & 173 & 0 \\
        Task 2b & Same origami sequence
        & 3 & 4 & 173 & 0 \\
        Task 2c--Mix & Mixed interventions
        & 3 & 4 & 172 & 1 \\
        Task 2c--Point & Point intervention
        & 3 & 4 & 172 & 1 \\
        Task 2c--Type & Mountain/Valley intervention
        & 1 & 2 & 165 & 8 \\
        Task 2c--Layer & Layer intervention
        & 1 & 2 & 9 & 164 \\
        \midrule
        \multicolumn{4}{r}{\textbf{Total}}
        & \textbf{864} & --- \\
        \bottomrule
    \end{tabular}
    \caption{Construction and statistics of the Task 2 variants. ``Skipped'' is measured relative to the 173 eligible source transitions. Task 2a and Task 2b use every eligible transition, whereas Task 2c retains a transition only when the required number of valid and visually distinct counterfactual distractors can be generated.}
    \label{tab:task2-statistics}
\end{table}

Overall, Task 2 contains 864 multiple-choice instances and 3,108 candidate options. Task 2a and Task 2b contribute 173 instances each, while the four Task 2c variants contribute 518 instances in total.

\begin{figure}[htbp]
\centering
\includegraphics[width=0.72\textwidth, trim=0.cm 0.cm 0.cm 0.cm]{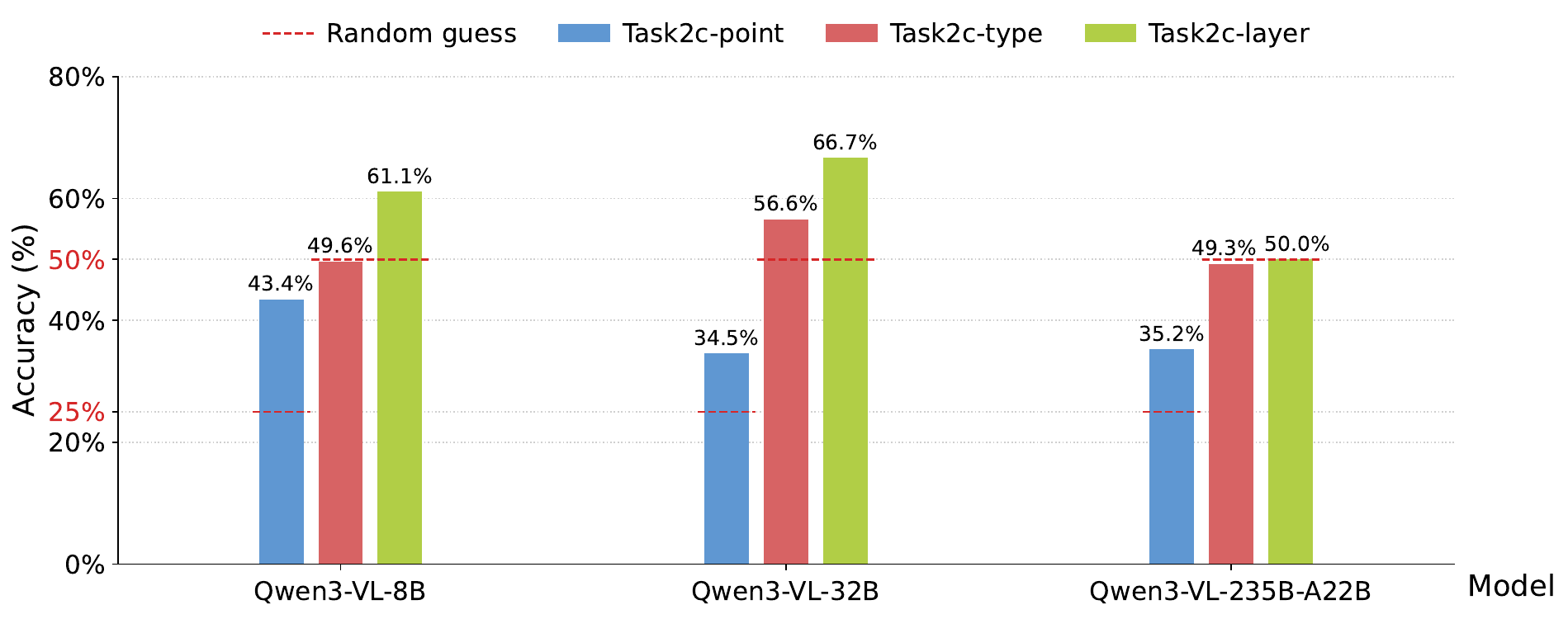}
\caption{\textbf{Accuracy across individual intervention types in Task 2c (Programmatic Grounding).} Point, Type, and Layer modify the fold-line points, mountain/valley type, and layer selection, respectively. The dashed lines mark the corresponding random-guess baselines (25\% for Point and 50\% for Type and Layer).}
\label{fig:task2c}
\end{figure}

{\subsection{Prompt}
\label{appendix:detail_experiment}}
\begin{tcolorbox}[breakable, title=Prompt for Task 2 ]

You are an expert in origami geometry and spatial reasoning.

You will receive candidate folded states (Options \{options\_csv\}), one Origin folded state, and one true program expression from the underlying action language.
Choose the option that is the exact state produced by applying that expression to the Origin.

IMPORTANT: Your final line must be exactly \texttt{\textbackslash boxed\{X\}}, where X is one of \{options\_or\}.
The evaluator parses the LAST boxed answer.

================================ \\
INFO ABOUT COORDINATE SYSTEM \\
================================

\begin{itemize}
    \item The original square sheet uses $0 \le x \le 3$ and $0 \le y \le 3$.
    \item Rendered images may be scaled, translated, or reframed, so DSL coordinates are not pixel coordinates.
    \item Use visible geometry (silhouette, vertices, edges, creases, flaps, and layer cues) to align coordinates with the rendered state.
\end{itemize}

================================ \\
TRUE PROGRAM EXPRESSION \\
================================

\begin{verbatim}
{expression}
\end{verbatim}

================================ \\
EXAMPLES (NO FORMAL GRAMMAR GIVEN) \\
================================

Before the target task, you will receive multimodal examples from the same language family.
Each example includes:
\begin{itemize}
    \item one expression string,
    \item a Before state (front/back),
    \item an After state (front/back) produced by that expression.
\end{itemize}

No formal grammar is provided. Infer language behavior from these examples plus visual evidence.

================================ \\
DECISION PROCEDURE \\
================================

\begin{enumerate}
    \item Read the true expression and infer the intended geometric action.
    \item Inspect both front and back views of Origin.
    \item Mentally apply one action step implied by the expression.
    \item Compare the predicted result against all options (\{options\_csv\}) on both views.
    \item Pick the single best-matching option.
\end{enumerate}

Do not rely on option order or superficial visual similarity alone.

================================ \\
INPUT ORDER \\
================================

Inputs appear in this order:
\begin{enumerate}
    \item Example block(s): expression + Before (front/back) + After (front/back).
    \item The target task: each valid option in \{options\_csv\} (front then back), each labeled by letter.
    \item Origin (front then back).
\end{enumerate}

Only options listed in \{options\_csv\} are valid.

================================ \\
OUTPUT CONTRACT \\
================================

You may reason briefly, but end with exactly one boxed answer line:

\[
\boxed{X}
\]

\end{tcolorbox}

\subsection{Additional Results}
\label{app:task2_perturbations}

We further break down Task 2c by the type of intervention applied to the program expression. These results reveal substantial differences in the difficulty of grounding different parts of the symbolic structure. Interventions to the \textbf{fold type} and \textbf{layer selection} are comparatively easier to detect. For fold-type interventions, Qwen3-VL-8B, Qwen3-VL-32B, and Qwen3-VL-235B-A22B achieve 49.6\%, 56.6\%, and 49.3\%, respectively, while layer interventions yield 61.1\%, 66.7\%, and 50.0\%. In contrast, \textbf{point interventions} are substantially harder, with accuracies of 43.4\%, 34.5\%, and 35.2\% for the three models. The \textbf{mixed intervention} condition is harder still, yielding 29.4\%, 35.4\%, and 34.1\%, respectively. These results suggest that models are better able to recognize relatively coarse symbolic changes, such as changes in fold type or layer selection, than fine-grained geometric changes to fold-line points. The mixed condition further indicates that combining multiple intervention types makes it difficult for models to identify the underlying symbolic change. Notably, model scale does not consistently improve performance across intervention types, particularly for the fine-grained interventions.

\end{document}